\documentclass[lettersize,journal]{IEEEtran}
\usepackage{amsmath,amsfonts}
\usepackage{algorithmic}
\usepackage{algorithm}
\usepackage{array}
\usepackage[caption=false,font=normalsize,labelfont=sf,textfont=sf]{subfig}
\usepackage{textcomp}
\usepackage{stfloats}
\usepackage{url}
\usepackage{verbatim}
\usepackage{graphicx}
\usepackage{cite}
\usepackage{amsmath} 
\usepackage{booktabs}
\usepackage{tabularx}
\usepackage{multirow}
\usepackage{makecell}
\usepackage{float}
\usepackage[dvipsnames]{xcolor}

\begin{document}

\title{Cross-Domain Few-Shot Segmentation via Ordinary Differential Equations over Time Intervals}

\author{
Huan Ni,
Qingshan Liu,~\IEEEmembership{Senior~Member,~IEEE}, 
Xiaonan Niu,
Danfeng Hong,~\IEEEmembership{Senior~Member,~IEEE},
Lingli Zhao,
Haiyan Guan,~\IEEEmembership{Senior~Member,~IEEE}

\thanks{This work was supported by the foundation from Beijing Key Laboratory of Advanced Optical Remote Sensing Technology under Grant AORS202310, and the Natural Science Foundation of
China under Grant U21B2044 and Grant 41801384. \it{(Corresponding author: Qingshan Liu; Xiaonan Niu.)}}
\thanks{Huan Ni and Haiyan Guan are with the School of Remote Sensing \& Geomatics Engineering, Nanjing University of Information Science \& Technology, Nanjing 210044, China. They are also with the Tiandu-Nuist Deep Sapce Exploartion Laboratory, Nanjing 210044, China (e-mail: nih2015@yeah.net; nih@nuist.edu.cn; guanhy.nj@nuist.edu.cn).}
\thanks{Qingshan Liu is with the School of Computer Science, Nanjing University of Posts and Telecommunications, Nanjing 210023, China (e-mail: qsliu@njupt.edu.cn).}
\thanks{Xiaonan Niu is with Nanjing Center, China Geological Survey, Nanjing 210016, China (niuxiaonan@mail.cgs.gov.cn).}
\thanks{Danfeng Hong is with the School of Automation, Southeast University, Nanjing 210096, China (e-mail: danfeng.hong@seu.edu.cn).}
\thanks{Lingli Zhao is with the School of Remote Sensing and Information Engineering, Wuhan University, Wuhan 430079, China (e-mail: zhaolingli@whu.edu.cn).}
}

\markboth{Journal of \LaTeX\ Class Files,~Vol.~14, No.~8, August~2021}%
{Shell \MakeLowercase{\textit{et al.}}: A Sample Article Using IEEEtran.cls for IEEE Journals}


\maketitle

\begin{abstract}
Cross-domain few-shot segmentation (CD-FSS) aims to segment unseen categories with very limited samples while alleviating the negative effects of domain shift between the source and target domains. At present, existing CD-FSS studies typically rely on multiple independent modules to enhance cross-domain adaptability. However, the independence among these modules hinders the effective flow of knowledge, making it difficult to fully leverage their collective potential. In contrast, this paper proposes an all-in-one module based on ordinary differential equations (ODEs) and the Fourier transform, resulting in a structurally concise method—Few-Shot Segmentation over Time Intervals (FSS-TIs). FSS-TIs not only explores a domain-agnostic feature space, but also achieves significant performance improvement through target-domain fine-tuning with extremely limited support samples. Specifically, the ODE modeling process incorporates nonlinear transformations and random perturbations of the amplitude and phase spectra, effectively simulating potential target-domain data distributions. Meanwhile, the analytical solution of the ODE is transformed into a theoretically infinitely iterable feature refinement process, thereby enhancing the learning capability under limited support samples. In this way, both the exploration of domain-agnostic features and the few-shot learning problem can be addressed through the optimization of the intrinsic parameters of the ODE. Moreover, during target-domain fine-tuning, we strictly constrain the support samples to match the settings of real-world CD-FSS tasks, without incurring additional annotation costs. For evaluation, we introduce nine datasets from substantially different domains and define sixteen CD-FSS tasks to comprehensively analyze the performance of FSS-TIs. The experimental results demonstrate the superiority of FSS-TIs over existing CD-FSS methods, and in-depth ablation studies further validate the cross-domain adaptability of FSS-TIs. 

\end{abstract}


\section{Introduction}
\label{SEC_INTRODUCTION}
\IEEEPARstart{B}{ased} on the concept of "episodes," few-shot segmentation (FSS) recognizes novel categories using very limited support exemplars, typically by transferring category-agnostic knowledge learned from abundant base categories to novel categories \cite{CDFSS-Nie2024, FSS-Liu2026}. In recent years, a series of representative FSS methods such as Prior Guided Feature Enrichment Network (PFENet) \cite{FSS-PFENET2022}, PFENet++ \cite{FSS-PFENET2024}, Holistic Prototype Activation (HPA) \cite{FSS-HPA2023}, and Base and Meta (BAM) \cite{FSS-BAM2023} have been proposed, illustrating that FSS is feasible and effective in reducing the annotation cost of pixel-level dense prediction for novel categories. Although FSS uses only very limited support exemplars for novel categories, it still needs many training samples for base categories \cite{CDFSS-Lei2022}. When the target domain including the novel categories is extremely scarce in resources and it is impossible to obtain several training samples of some base categories, we must use the training samples of the base categories in the resource-rich domain (source domain) to train an FSS model \cite{CDFSS-Tong2025b}. In this case, domain shifts \cite{OTHER-Chen2024} between the training samples of base categories and the support exemplars of novel categories emerge. Unfortunately, existing FSS methods often exhibit subpar performance when confronted with domain shifts \cite{CDFSS-Su2024, CDFSS-Fan2025}. 

To address this domain shift issue in FSS tasks, cross-domain few-shot segmentation (CD-FSS) is proposed. There are two fundamental challenges of CD-FSS. On the one hand, there are domain shifts between the source and target domains. The base categories are from the source domain, and the novel categories are from the target domain \cite{CDFSS-Liu2025}. On the other hand, the support exemplars of novel categories in the target domain is so scarce that the fine-tuning or distribution alignment is very challenging \cite{CDFSS-Wang2022, CDFSS-Tong2025a}. Correspondingly, existing approaches can be roughly categorized into three groups. The first type of method operates solely on the source domain. These approaches aim to learn domain-agnostic feature representations from the training samples of base categories \cite{CDFSS-Chen2024a} or simulate diverse domain styles through feature perturbation \cite{CDFSS-Su2024} to enhance the model's adaptability to novel categories in the target domain. The second type of method focuses on the target domain. By introducing additional layers or augmenting the input image views \cite{CDFSS-Herzog2024}, the method alleviates the overfitting problem caused by the limited support samples of novel categories in the target domain. The third type of method jointly addresses both the source and target domains. These methods simultaneously improve the generalization capability of features extracted from base categories in the source domain and perform fine-tuning on the target domain \cite{CDFSS-Chen2024b, CDFSS-Nie2024}.

From a practical standpoint, for the first type of methods, since identifying each novel category in the target domain always requires one or a few support samples, why can’t these support samples also be used for target-domain fine-tuning? Moreover, this few-shot fine-tuning process is simple and time efficient. For the second and third types of methods, the current way that target-domain fine-tuning uses the support samples of novel categories does not align with the requirements of real-world tasks. In practical FSS tasks, to reduce the cost of manual annotation, for a $K$-shot task, we usually provide exactly $K$ support samples for each category. Consequently, during target-domain fine-tuning, the support samples for novel categories can only be drawn from the already-provided $K$ support samples. During the accuracy evaluation process, accuracy must not be calculated for images already selected as support samples; otherwise, the evaluation results will be artificially inflated. However, in most CD-FSS methods, in each training epoch of target-domain fine-tuning, $K$ samples are randomly selected for each novel category from the entire dataset. If the training runs for $20$ epochs, the number of samples used for each novel category in fine-tuning will far exceed $K$, which is inconsistent with real-world conditions. Furthermore, during accuracy evaluation, the support samples that were randomly selected for fine-tuning are not excluded, leading to reported testing accuracy being higher than what would be obtained in practical applications.

From a methodological perspective, the studies in \cite{CDFSS-Nie2024} and \cite{CDFSS-Chen2024b} are highly representative. The work in \cite{CDFSS-Nie2024} systematically demonstrates the importance of target-domain fine-tuning. Furthermore, this method extends the prediction paradigm of FSS and incorporates this extension into an iterative framework. This iterative framework can be perceived as increasing the training depth of the model prediction process under the condition of limited novel category samples, thereby enhancing the optimization of parameters. The study of \cite{CDFSS-Chen2024b} transforms domain-specific features into a domain-agnostic feature space during the training process on base categories. Moreover, this method establishes self-matching and dual hyper-correlation mechanisms within the domain-agnostic feature space, which not only deepens the feature representation but also prevents overfitting when only a few samples are available in the target domain. These two methods have improved the performance of CD-FSS, but there are still some issues that need further improvement. First, the study in \cite{CDFSS-Nie2024} does not explore domain-agnostic feature representations and lacks a theoretical explanation. Second, the method proposed by \cite{CDFSS-Chen2024b} is structurally complex. The fundamental issue lies in the separation between the domain-agnostic feature space's exploration and the feature representation's deepening processes. This independent strategy not only complicates the model architecture but also impedes the knowledge interaction between the two processes. In contrast, we have been committed to developing methods that are theoretically grounded and structurally concise, aiming to augment CD-FSS performance while maintaining extensibility and practical applicability. 

In recent years, many effective networks, such as ResNet \cite{He-2016} and PolyNet \cite{OTHER-PolyNet2017}, can be interpreted as different numerical discretizations of differential equations \cite{ODEs-Lu2018}. This perspective has played an inspirational role in the development of the method proposed in this paper. However, unlike these methods, we do not directly discretize the ordinary differential equation (ODE). Instead, after obtaining the analytical solution to the initial value problem (IVP) of the ODE, we embed this solution into the deep learning framework using numerical computation methods. We find that the numerical form of this solution over time intervals naturally integrates the processes of discovering domain-agnostic feature spaces and deepening feature representations, thereby improving CD-FSS performance while maintaining a concise model architecture.

Specifically, this paper introduces the Fast Fourier Transform to decompose features into amplitude and phase spectra, handling them separately to transform global texture information and fine structural information, respectively. An ODE relationship with learnable parameters is then assumed to exist between the spectra of domain-specific and domain-agnostic features. The analytical solution of the ODE is derived, and numerical approximation with random spectral perturbations is performed over a sequence of tiny time intervals, forming a learning process that iteratively optimizes feature representation. With this enhanced feature representation, the CD-FSS performance is substantially improved. The main contributions of this paper are as follows: 
\begin{itemize}
\item{Based on FFT and ODEs, this paper develops an interpretable all-in-one module for CD-FSS, termed Transformation over Time Intervals (TTIs). The modeling process of TTIs incorporates nonlinear transformations and random perturbations to convert domain-specific amplitude and phase spectra into instantaneously alignable states, and further establishes a time-varying dynamical relationship between these states and the domain-agnostic spectra.} 
\item{We transform the analytical solution of the ODE into an iterative form over a sequence of small time intervals to deepen feature representations. Furthermore, via numerical computations, the solution of the ODE is embedded into the deep learning framework, forming a feature learning process that can be theoretically iterated indefinitely for progressive refinement.} 
\item{This study designs a target-domain fine-tuning procedure with strict constraints of support samples, which is more consistent with real-world settings. The procedure restricts the sampling scope of support samples for novel categories in the target domain and ensures their complete separation from the data used for accuracy evaluation.}
\end{itemize}

When the third innovation is combined with TTIs, it yields a novel CD-FSS method, called few-shot segmentation over time intervals (FSS-TIs). To facilitate readers' understanding of the notation system used in this paper, we present the naming logic of our symbols as follows. First, we use the initial letter of a noun to denote the corresponding variable in the algorithm, e.g., feature $\rightarrow f$. Second, superscripts indicate the attributes of a variable, while subscripts denote indices. For example, the source domain is denoted as $\mathcal{D}^{s}$, where $\mathcal{D}\rightarrow \text{domain}$ and $s\rightarrow \text{source}$. In the time step $t_n$, $t\rightarrow \text{time}$ and $n$ denotes the index of the time step. Third, since CD-FSS involves multiple concepts, a single superscript is often insufficient to fully describe the attributes of a variable or distinguish it from others. Therefore, we sometimes use multiple superscripts. For instance, the $k$-th support image from the base category $c^s$ in the source domain is denoted as $I_k^{s, s, c^s}$, where the first $s \rightarrow \text{source}$, the second $s \rightarrow \text{support}$, and $c^s \rightarrow \text{category } c^s$, respectively.

\section{Related Works}
\label{SEC_RELATEDWORKS}
The foundation of CD-FSS lies in FSS. This section first introduces FSS, followed by a discussion of the current research progress in CD-FSS. 

\subsection{Few-shot Segmentation}
\label{SEC_R_FSS}
FSS places the general semantic segmentation in a few-shot scenario, where models perform dense pixel labeling on novel categories with only a few support samples \cite{FSS-PFENET2022}. More specifically, FSS reorganizes the training and testing sets into a series of episodes \cite{EPISODE-Guo2025}. In the training episodes, there are only the base categories $C_c^b\{c_i^b, i=1, \cdots, c^b\}$, and in the testing episodes, there are only the novel categories $C_c^n\{c_i^n, i=1, \cdots, c^n\}$, and $C_c^b \cap C_c^n = \emptyset $ \cite{FSS-Bi-2025}. Each episode $\mathcal{E}=\{\mathcal{S}, \mathcal{Q}\}$ consists of a small support set $\mathcal{S}=\{(x_k^s, m_k^s)\}_{k=1}^K$ and a query set $\mathcal{Q}=\{x^q, m^q\}$ \cite{FSS-BAM2023}, where $x_k^s$ is the $k^{th}$ support image and $m_k^s$ is the corresponding binary mask of $x_k^s$. $x^q$ is the query image, and $m^q$ is the corresponding binary mask of $x^q$ \cite{FSS-HPA2023}. According to the settings of $K$, the FSS task is named a $K$-shot task. In general, FSS studies mainly focus on tasks under the $1$-shot and $5$-shot settings \cite{FSS-Bi-2025}. Once the model is trained using the episodes on the base categories, the model is directly applied in the testing episodes using only very limited support samples of novel categories without further optimization \cite{FSS-Chen-2024}. Existing FSS methods can be divided into (1) prototype matching-based, (2) feature fusion-based, and (3) pixel matching-based methods \cite{FSS-Shaban-2017, FSS-Bi-2025}. 

Prototype matching-based methods introduce masked average pooling \cite{FSS-SG-One-2020}, normalized masked average pooling \cite{FSS-AMP-2019}, or multi-frequency pooling \cite{FSS-Wen-2024} to generate a prototype or several prototypes for each category (represented as feature vectors) based on the support image features \cite{FSS-Dong-2018}. Subsequently, pixel-level dense prediction for the target category is achieved by matching the prototypes of the target category with query image features using techniques such as prototype alignment regularization \cite{FSS-PANet-2019}, cosine similarity \cite{FSS-Nguyen-2019}, cross-reference mechanism \cite{FSS-CRNet-2020}, and self-support matching \cite{FSS-SSP-2022}. To generate multiple prototypes for each class, the $K$-means clustering \cite{FSS-Liu-2020}, prototype mixture models \cite{FSS-PMMs-2020} and dividing mechanism \cite{FSS-Lang-2024} are introduced to transform support image features into multiple prototypes with diverse attributes. Feature fusion-based methods combine query and support features or fuse more useful information into query and support features to make more reliable predictions. To combine the query and support features, prior masks \cite{FSS-PFENET2022}, complementary feature learning manners \cite{FSS-APANet-2023}, Intermediate Prototype Mining Transformer (IPMT) \cite{FSS-Liu-2022a}, the cross-attention and self-attention mechanisms \cite{FSS-FECANet-2023}, and intermediate prototypes \cite{FSS-Liu2025} are introduced. To further fuse more useful information, 4D convolution-based multi-scale feature fusion \cite{FSS-Min-2021}, Non-Target Region Elimination (NTRE) \cite{FSS-Liu-2022b}, Cross-Referenced Decoder (CRD) \cite{FSS-HPA2023}, conditional Gaussian enhancement \cite{FSS-Ni-2025}, and the prompt-driven scheme  “Prompt and Transfer” \cite{FSS-Bi-2025} are proposed. Subsequently, the textual cues \cite{FSS-Nandam2025} and large language models (LLMs) \cite{FSS-Zhu-2025, FSS-Karimi2025} are applied. Pixel-matching-based methods have benefited from recent advances in computing hardware. These methods enable the extension of one or several prototypes to the pixel level via agent matching decoder \cite{FSS-Wang-2022}, cycle-consistent attention mechanism \cite{FSS-Zhang-2021}, scaled-cosine (SC) mechanism \cite{FSS-Xu-2023}, and hierarchical dense correlation distillation \cite{FSS-Peng-2023} for a more comprehensive representation of features.

\subsection{Cross-domain Few-shot Segmentation}
\label{SEC_R_CDFSS}
In real-world scenarios, domain shifts often exist between the training samples (from the source domain) and the data to be predicted (from the target domain) due to differences in data distribution \cite{OTHER-Xu2025}, which has led to the emergence of CD-FSS \cite{CDFSS-Lei2022}. At present, existing CD-FSS approaches can be roughly categorized into three types: methods that operate only on the source domain, methods that focus solely on the target domain, and methods that operate on both the source and target domains. 

The first type focuses on designing the learning process for base categories in the source domain. Some studies assume the existence of a domain-agnostic feature space and introduce techniques during model training such as anchor point mapping based on transformation matrices \cite{CDFSS-Huang2023, CDFSS-He2024}, bidirectional 3D convolutions \cite{CDFSS-Chen2024a}, feature disentanglement \cite{CDFSS-Chen2024c}, the pre-trained foundation model Segment Anything \cite{OTHER-SAM2023}, and prototype alignment between the support and query samples \cite{CDFSS-Lu2022}. These techniques aim to establish a mapping from domain-specific feature space to the domain-agnostic one. During inference, it is assumed that this learned mapping remains valid for the target domain, allowing the transformed target-domain features to be used for predicting novel categories. Additionally, without accessing the target domain, the study in \cite{CDFSS-Su2024} attempts to simulate diverse target-domain styles by perturbing the statistical properties of source-domain features, which helps to reduce the impact of domain shifts. The second type \cite{CDFSS-Herzog2024} adopts a more radical approach by entirely removing the training and modeling process related to the source domain, focusing exclusively on the target domain. This method uses a pre-trained deep learning model as the backbone for feature extraction and completely discards source-domain training. It still adopts an episodic learning structure and alleviates overfitting under limited target-domain support by adding auxiliary layers and enhancing the input image views. The third type emphasizes both source-domain feature representation and target-domain fine-tuning. Some studies \cite{CDFSS-Chen2024b, CDFSS-Lei2022} also assume the existence of a domain-agnostic feature space. Once this space is identified, parameter fine-tuning is performed using a few target-domain novel-category samples \cite{CDFSS-Nie2024, CDFSS-Li2025}. Alternatively, the approach \cite{CDFSS-Wang2022} learns distribution-specific features in the source domain and stores them in a “meta-memory bank”, and then applies to enhance target-domain features. 

These methods have improved the CD-FSS capability to some extent. However, most of them still remain at the level of network architecture design and lack the ability to explicitly establish interpretable mappings between domain-specific and domain-agnostic feature spaces. 

In addition, we find that cross-domain approaches based on frequency transformation possess mathematical and physical interpretability. Among them, Fourier Domain Adaptation (FDA) \cite{FFT-FDA-2020} replaces the low-frequency part of the amplitude spectrum of the source-domain image with the corresponding part from the target-domain image, so as to alleviate the distribution shift between different domains. This idea was later extended by Class Aware Frequency Transformation (CAFT) \cite{FFT-CAFT-2023} to the class-wise feature space. However, the basic assumption of the source-domain training process in CD-FSS is that the model cannot access target-domain images, which makes such cross-domain ideas difficult to apply in this field. Meanwhile, these interpretable cross-domain ideas are highly inspiring, leading to CD-FSS methods based on frequency transformation, such as Amplitude-Phase Masker (APM) \cite{CDFSS-Tong2024}. This method introduces two adaptive mask parameters to filter out the parts of phase and amplitude spectra that are unfavorable for domain adaptation. While partially adopting the aforementioned frequency transformation strategy, this paper further takes the periodicity of the phase spectrum into consideration and proposes an ODE transformation method for both the phase and amplitude spectra, which provides an interpretable dynamical system for exploring the domain-agnostic feature space.

\section{FSS-TIs for CD-FSS}
\label{SEC_FSSTISORCDFSS}
\subsection{Problem Formulation}
\label{SEC_PF}
CD-FSS assumes that the training samples of base categories belong to the source domain, while the test samples of novel categories belong to the target domain, with a domain shift existing between the source and target domains. Let $\mathcal{D}^{s}=\{\mathcal{X}^s, \mathcal{Y}^s\}$ and $\mathcal{D}^{t}=\{\mathcal{X}^t, \mathcal{Y}^t\}$ are the source and target domains, where $\mathcal{X}^s$ and $\mathcal{X}^t$ are the data distribution, and $\mathcal{Y}^s$ and $\mathcal{Y}^t$ are the label space. The presence of a domain shift in CD-FSS tasks implies that $\mathcal{X}^s \neq \mathcal{X}^t$ and $\mathcal{Y}^s \cap \mathcal{Y}^t = \emptyset$. When considering the episode training strategy in FSS, the samples belonging to each category in $\mathcal{D}^{s}$ and $\mathcal{D}^{t}$ are reorganized into a range of episodes $\{\mathcal{E}_i^{s, c^s}=\{S_i^{s, c^s}, Q_i^{s, c^s}\}, i=1,\cdots, N^{s, c^s}\}$ and $\{\mathcal{E}_j^{t, c^t}=\{S_j^{t, c^t}, Q_j^{t, c^t}\}, j=1,\cdots, N^{t, c^t}\}$, where $N^{s, c^s}$ is the number of episodes for the $c^s$ category in $\mathcal{D}^{s}$ and $N^{t, c^t}$ is that for the $c^t$ category in $\mathcal{D}^{t}$. Taking an episode $\mathcal{E}_i^{s, c^s}=\{S_i^{s, c^s}, Q_i^{s, c^s}\}$ from the source domain as an example, $S_i^{s, c^s}=\{I_k^{s, s, c^s}, M_k^{s, s, c^s}\}_{k=1}^K$ is the support set, and $Q_i^{s, c^s}=\{I^{s, q, c^s}, M^{s, q, c^s}\}$ is the query set, where $I_k^{s, s, c^s}$ and $I^{s, q, c^s}$ are the support and query images, and $M_k^{s, s, c^s}$ and $M^{s, q, c^s}$ are their masks (ground truth) for the $c^s$ category. $K$ has the same meaning as that in FSS settings, as illustrated in \ref{SEC_R_FSS}. 

When adopting the CD-FSS approach that operates on both the source and target domains, the overall framework is divided into three steps \cite{CDFSS-Nie2024}:
\begin{itemize}
\item{Training the model in $\mathcal{E}_i^{s, c^s}=\{S_i^{s, c^s}, Q_i^{s, c^s}\}$.}
\item{Fine tuning the trained model based on a very limited number of $\mathcal{E}_j^{t, c^t}=\{S_j^{t, c^t}, Q_j^{t, c^t}\}$.}
\item{Testing the adapted model in $\mathcal{D}^{t}$.}
\end{itemize}

\subsection{Method Overview}
\label{SEC_MO}

\begin{figure}[!t]
\centering
\includegraphics[width=3.5in]{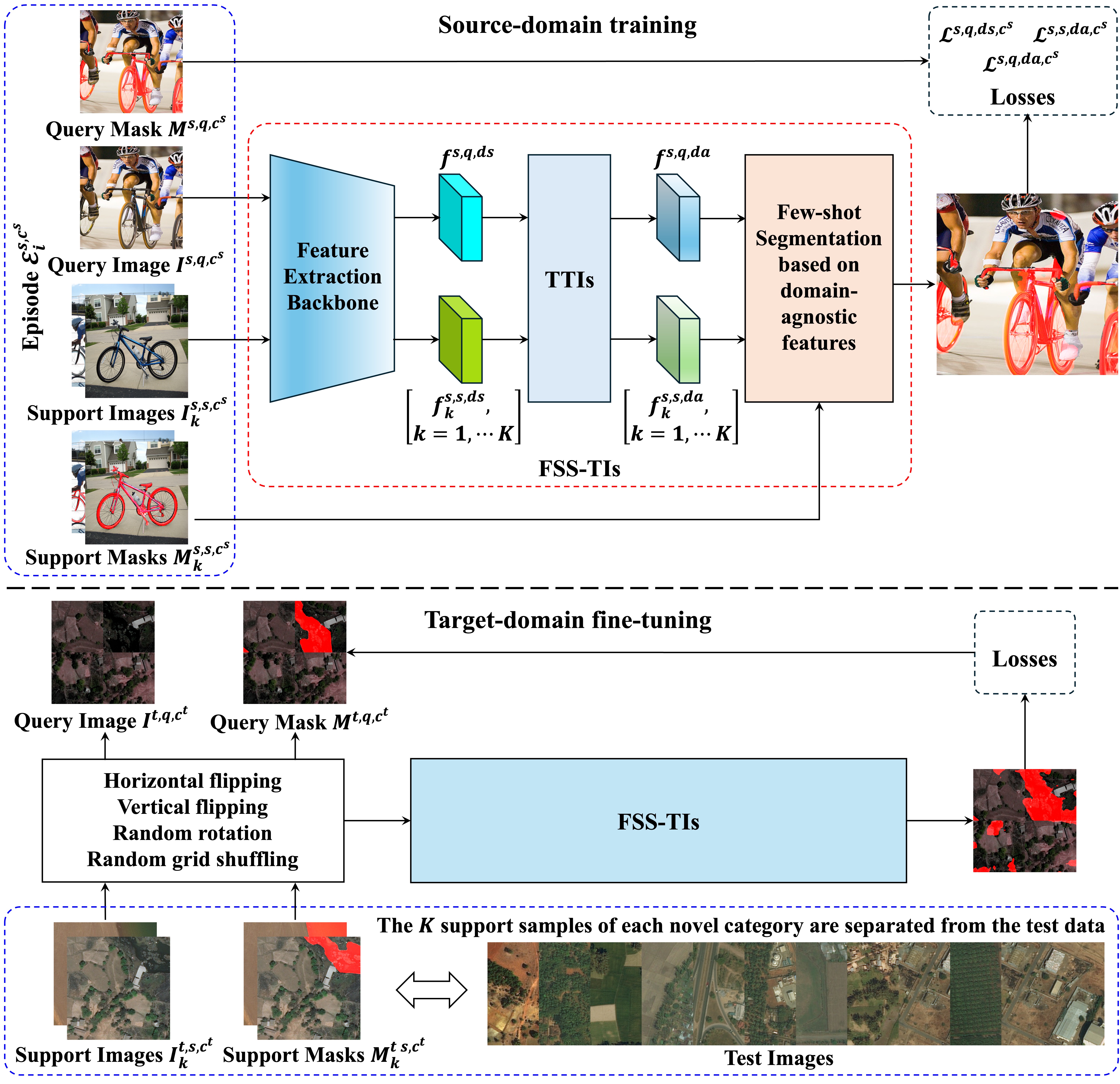}
\caption{The overall data flow of the proposed method FSS-TIs. The data flow includes two parts: source-domain training and target-domain fine-tuning. The all-in-one module TTIs serves as the core of FSS-TIs.} 
\label{FIG_OVERVIEW}
\end{figure}

Fig. \ref{FIG_OVERVIEW} illustrates the overall data flow of FSS-TIs, which consist of two parts: source-domain training and target-domain fine-tuning. At the center of this framework is the all-in-one module TTIs, which serves as the core of the method. As can be observed, the overall data flow of FSS-TIs is concise, and the model's architecture is equally streamlined, fully aligning with our original design intention. 

During the source-domain training process, we use a randomly sampled episode from a single training loop, denoted $\mathcal{E}_i^{s, c^s}=\{S_i^{s, c^s}, Q_i^{s, c^s}\}$, as an example to illustrate the data flow. Given the support set $S_i^{s, c^s}=\{I_k^{s, s, c^s}, M_k^{s, s, c^s}\}_{k=1}^K$, which contains $K$ support images and their corresponding masks, and the query set $Q_i^{s, c^s}=\{I^{s, q, c^s}, M^{s, q, c^s}\}$, which contains the query image and its mask, we first input all $K$ support images ${[I_k^{s, s, c^s}, k=1, \cdots, K]}$ together with the query image $I^{s, q, c^s}$ into the feature extraction backbone. This produces the domain-specific query feature $f^{s, q, ds}$ and a set of domain-specific support features ${[f_k^{s, s, ds}, k=1, \cdots, K]}$. These features are then fed into the TTIs module, which transforms them into domain-agnostic features, namely, $f^{s, q, da}$ for the query image and ${[f_k^{s, s, da}, k=1, \cdots, K]}$ for the support images. Based on these domain-agnostic features, FSS is performed on the query image $I^{s, q, c^s}$. During backpropagation, corresponding loss functions are designed according to the properties of TTIs, enabling the optimization and update of both the feature extraction backbone and the built-in parameters of TTIs.

In the target-domain fine-tuning process, we strictly restrict the use of support samples. Specifically, in a $K$-shot task, target-domain fine-tuning for each novel category is conducted using only $K$ support samples. If $K=1$, then only one support sample is used for the fine-tuning of each novel category. Moreover, the support samples used for fine-tuning must strictly match those used in the target domain's FSS testing. This constraint is equivalent to the fine-tuning part illustrated in Figure \ref{FIG_OVERVIEW}, where the $K$ support samples for each novel category are kept completely separate from the test data. Only under this condition can we ensure that no additional annotation cost is introduced to practical tasks. Furthermore, since this is a FSS task, parameter updates during the fine-tuning process also rely on query images and their ground-truth masks. To avoid extra manual annotation, this study randomly applies flipping, rotation, and grid shuffling to the support images of each novel category during the fine-tuning training loop, thereby generating corresponding query images and masks from the support images in an automatic manner.

\subsection{TTIs}
As displayed in Fig. \ref{FIG_OVERVIEW}, in both the source-domain training and target-domain fine-tuning, the domain-specific features of the support and query images are sequentially transformed by TTIs. To avoid overwhelming readers with excessive symbols, in this section we use $f^{ds}$ to denote any of the above domain-specific features and $f^{da}$ to denote the domain-agnostic feature obtained by transforming $f^{ds}$ through TTIs. We can regard $f^{ds}$ and $f^{da}$ as functions of time, and thus represent them as $f^{ds}(t)$ and $f^{da}(t)$, respectively.

\subsubsection{Fundamental Assumptions}
\label{SEC_FA}

Here, we assume the existence of a domain-agnostic feature space and further assume that the transformation from the domain-specific feature space to the domain-agnostic one can be modeled using the Fourier Transform and ODEs. Let the amplitude spectrum $\mathcal{A}(t)$ and the phase spectrum $\mathcal{P}(t)$ obtained from $f^{ds}(t)$ via the Fast Fourier Transform (FFT) be defined as follows:
\begin{equation}
\label{M_EQ1}
\mathcal{A}(t) e^{\mathcal{P}(t)} = \text{FFT}(f^{ds}(t)). 
\end{equation}
The amplitude spectrum $\mathcal{A}(t)$ mainly captures the overall information, and the phase spectrum $\mathcal{P}(t)$ mainly presents the shape, edge, and structures. These two types of information possess distinctly different characteristics. They must be processed separately to be transformed into the domain-agnostic spectra $\widehat{\mathcal{A}}(T)$ and $\widehat{\mathcal{P}}(T)$. Subsequently, the domain-agnostic feature $f^{da}$ is obtained by performing the inverse Fourier transform: 
\begin{equation}
\label{M_EQ2}
f^{da} = \text{IFFT}(\widehat{\mathcal{A}}(T) e^{i\widehat{\mathcal{P}}(T)}),
\end{equation}

Now, the most critical issue is how the domain-specific spectra are transformed into domain-agnostic spectra over the time interval $[t_0, T]$. For the domain-specific spectra $\mathcal{A}(t)$ and $\mathcal{P}(t)$, we assume that there exist instantaneous alignable states $q^a(t)$ and $q^p(t)$, respectively, which are obtained from the domain-specific spectra or the previous states of the domain-agnostic spectra ($\ddot{\mathcal{A}}$ and $\ddot{\mathcal{P}}$) on $\ddot{t}$ through nonlinear transformations with learnable parameters, as follows: 
\begin{equation}
\label{M_EQ3}
q^a(t) = \ddot{\mathcal{A}}(\ddot{t}) * (0.5 + \text{Sigmoid}(\text{Convs}(\ddot{\mathcal{A}}(\ddot{t})))),
\end{equation}

\begin{equation}
\label{M_EQ4}
\begin{cases}
{\begin{bmatrix} \text{sin}(\ddot{\mathcal{P}}(t))\\ \text{cos}(\ddot{\mathcal{P}}(t)) \end{bmatrix}} = \text{Convs}({\begin{bmatrix} \text{sin}(\ddot{\mathcal{P}}(\ddot{t}))\\ \text{cos}(\ddot{\mathcal{P}}(\ddot{t})) \end{bmatrix}}), \\
q^p(t) = \text{arctan} \frac{\text{sin}(\ddot{\mathcal{P}}(t)) / \sqrt{\text{sin}^2 (\ddot{\mathcal{P}}(t)) + \text{cos}^2 (\ddot{\mathcal{P}}(t))}}{\text{cos}(\ddot{\mathcal{P}}(t)) / \sqrt{\text{sin}^2 (\ddot{\mathcal{P}}(t)) + \text{cos}^2 (\ddot{\mathcal{P}}(t))}}. 
\end{cases}
\end{equation}

Eq.~(\ref{M_EQ3}) adaptively adjusts the overall style of features by transforming the amplitude spectrum, where $\text{Sigmoid}(\cdot)$ denotes the Sigmoid activation function, which constrains the convolutional responses to the range of $[0,1]$. The constant $0.5$ enables these responses to both suppress and enhance the original amplitude spectrum in a spatially adaptive manner. Eq.~(\ref{M_EQ4}) simulates structural deformation under extreme domain shifts by transforming the phase spectrum. Specifically, the phase is first transformed into sine and cosine values within the range of $[-1,1]$, and then converted back into the phase representation after the transformation process. The main reason for doing so is that the phase representation inherently contains periodic boundaries. For example, when the phase is represented using values within the interval $[0,2\pi]$, $0$ and $2\pi$ lie on the periodic boundary. Although they represent the same direction, they are numerically far apart. Various transformations defined in the Euclidean space, such as convolution, do not take this property of the phase into consideration. Therefore, directly applying such operations to the phase may introduce discontinuous jumps around the periodic boundary and lead to unpredictable instability. In contrast, sine and cosine representations transform the phase onto the unit circle in a two-dimensional Euclidean space, thereby naturally preserving the periodic continuity of the phase. Combined with the normalization process in the second row of Eq.~(\ref{M_EQ4}), the convolved sine and cosine values can be projected back onto the unit circle, thereby effectively avoiding the discontinuity problem around the phase boundary.

Based on the above alignable states $q^a(t)$ and $q^p(t)$, we define the following ODEs: 
\begin{equation}
\label{M_EQ5}
\begin{cases}
\frac{\text{d} \widehat{\mathcal{A}}(t)}{\text{d} t} = - \epsilon e^a(t) = - \epsilon (\widehat{\mathcal{A}}(t) - q^a(t)),\\
\frac{\text{d} \widehat{\mathcal{P}}(t)}{\text{d} t} = - \epsilon e^p(t) = - \epsilon (\widehat{\mathcal{P}}(t) - q^p(t)).
\end{cases}
\end{equation}

These ODEs define a stable error-contraction mechanism, which drives the domain-agnostic spectra $\widehat{\mathcal{A}}(t)$ and $\widehat{\mathcal{P}}(t)$ to evolve along the directions specified by $q^a(t)$ and $q^p(t)$. As a result, during the learning process of the internal parameters of $q^a(t)$ and $q^p(t)$, the domain-agnostic spectra can be derived. In other words, the initially input domain-specific spectra $\mathcal{A}(t)$ and $\mathcal{P}(t)$ in Eqs.~\ref{M_EQ3} and \ref{M_EQ4} are transformed into the domain-agnostic spectra $\widehat{\mathcal{A}}(t)$ and $\widehat{\mathcal{P}}(t)$. In addition, $\epsilon$ is a learnable parameter constrained to be positive. To ensure its positivity, we adopt a simple trick: 
\begin{equation}
\label{M_EQ6}
\epsilon = \ln (1 + e^{\ddot{\epsilon}})
\end{equation}
Here, $\ln(\cdot)$ is the natural logarithm. We let $\ddot{\epsilon}$ be the directly learnable parameter, and consequently $\epsilon$ is also learned from the data. The above dynamical system not only ensures modeling simplicity, but also admits a rigorous analytical solution. 

According to the general theory of ODEs, given $\widehat{\mathcal{A}}(t_0)$ and $\widehat{\mathcal{P}}(t_0)$ (where $t_0$ is the initial time), an IVP is formulated, and each equation in (\ref{M_EQ5}) has a unique solution:  
\begin{equation}
\label{M_EQ7}
\begin{cases}
\widehat{\mathcal{A}}(T) = e^{-\int_{t_0}^T \epsilon\ \text{d}r} (\widehat{\mathcal{A}}(t_0) + \epsilon \int_{t_0}^T e^{\int_{t_0}^s \epsilon \ \text{d}v} q^a(s)\text{d}s) ,\\
\widehat{\mathcal{P}}(T) = e^{-\int_{t_0}^T \epsilon\ \text{d}r} (\widehat{\mathcal{P}}(t_0) + \epsilon \int_{t_0}^T e^{\int_{t_0}^s \epsilon \ \text{d}v} q^p(s)\text{d}s).
\end{cases}
\end{equation}

\subsubsection{Numerical Computation of ODEs}
\label{SEC_NCODE}

Although the ODEs in Eq.~(\ref{M_EQ5}) have an analytical solution as shown in Eq.~(\ref{M_EQ7}), it can be observed that the analytical solution involves complex integrals and thus remains intractable for direct computation. To tackle this issue, there are generally two possible approaches. The first is to numerically compute the ODEs in Eq.~(\ref{M_EQ5}) directly, such as via the explicit Euler method (an iterative numerical computation method): 
\begin{equation}
\label{M_EQ8}
\begin{cases}
\widehat{\mathcal{A}}(t_{n+1}) =  \widehat{\mathcal{A}}(t_{n}) + h(-\epsilon(\widehat{\mathcal{A}}(t_{n}) - q_{t_n}^a)),\\
\widehat{\mathcal{P}}(t_{n+1}) =  \widehat{\mathcal{P}}(t_{n}) + h(-\epsilon(\widehat{\mathcal{P}}(t_{n}) - q_{t_n}^p)),
\end{cases}
\end{equation}
where $h = t_{n+1} - t_n$ is the time interval. 

The second approach is to start from the analytical solution of the ODEs and numerically compute the integrals involved in the analytical solution. The analytical solutions in Eq.~(\ref{M_EQ7}) can also be transformed into an iterative form similar to Eq.~(\ref{M_EQ8}). First, taking the first equation in Eq.~(\ref{M_EQ7}) as an example, it can be transformed into the following form: 
\begin{equation}
\label{M_EQ9}
\widehat{\mathcal{A}}(T) = e^{-\epsilon(T-t_0)} (\widehat{\mathcal{A}}(t_0) + \epsilon \int_{t_0}^T e^{\epsilon(s-t_0)} q^a(s)\text{d}s).
\end{equation}
To facilitate the derivation of the numerical solution in iterative form, let $T = t_{n+1}$. Then, according to the rules of integral operations, we obtain:
\begin{equation}
\label{M_EQ10}
\begin{split}
\widehat{\mathcal{A}}(t_{n+1}) = & e^{-\epsilon(t_{n+1}-t_0)} (\widehat{\mathcal{A}}(t_0) + \epsilon \int_{t_0}^{t_n} e^{\epsilon(s-t_0)} q^a(s)\text{d}s
\\ & + \epsilon \int_{t_n}^{t_{n+1}} e^{\epsilon(s-t_0)} q^a(s)\text{d}s).
\end{split}
\end{equation}
Furthermore, it can be observed that:
\begin{equation}
\label{M_EQ11}
\begin{split}
\widehat{\mathcal{A}}(t_{n}) = & e^{-\epsilon(t_{n}-t_0)} (\widehat{\mathcal{A}}(t_0) + \epsilon \int_{t_0}^{t_n} e^{\epsilon(s-t_0)} q^a(s)\text{d}s).
\end{split}
\end{equation}
It follows that:
\begin{equation}
\label{M_EQ12}
\begin{split}
\widehat{\mathcal{A}}(t_0) = & e^{\epsilon(t_n-t_0)}\widehat{\mathcal{A}}(t_n) - \epsilon \int_{t_0}^{t_n} e^{\epsilon(s-t_0)} q^a(s)\text{d}s. 
\end{split}
\end{equation}
Substituting Eq.~(\ref{M_EQ12}) back into Eq.~(\ref{M_EQ10}) and simplifying, we obtain the following analytical solution in iterative form:
\begin{equation}
\label{M_EQ13}
\begin{split}
\widehat{\mathcal{A}}(t_{n+1}) = & e^{-\epsilon(t_{n+1}-t_n)} \widehat{\mathcal{A}}(t_n)  + \epsilon \int_{t_n}^{t_{n+1}} e^{-\epsilon(t_{n+1}-s)} q^a(s)\text{d}s.
\end{split}
\end{equation}
Similarly, the second equation in Eq.~(\ref{M_EQ7}) can be transformed into the following iterative form:
\begin{equation}
\label{M_EQ14}
\begin{split}
\widehat{\mathcal{P}}(t_{n+1}) = & e^{-\epsilon(t_{n+1}-t_n)} \widehat{\mathcal{P}}(t_n)  + \epsilon \int_{t_n}^{t_{n+1}} e^{-\epsilon(t_{n+1}-s)} q^p(s)\text{d}s.
\end{split}
\end{equation}
In this iterative form, we only need to numerically compute the integral in the last term over the small time interval $h = t_{n+1} - t_n$. Compared with the numerical computation of the global analytical solution in Eq.~(\ref{M_EQ9}), the advantage of numerically computing this iterative-form solution is evident, since the truncation error of any numerical method decreases as $h$ becomes smaller. The question that follows is whether it is better to adopt the numerical computation process based on the iterative solution in Eqs.~(\ref{M_EQ13}, \ref{M_EQ14}) or to use the direct numerical methods for the ODEs in the first approach such as Eq.~\ref{M_EQ8}.

At present, there already exist a large number of numerical methods for integrals and ODEs. It is not wise to analyze, one by one, the various numerical integration methods for Eqs.~(\ref{M_EQ13}, \ref{M_EQ14}) as well as the truncation errors of every numerical method applicable to ODEs, because this would involve an enormous amount of work, while the final conclusion would still be that their truncation errors are generally comparable. However, this does not mean that the two types of approaches are equivalent for the ODEs presented in this paper. The key point is that the ODEs in Eq.~(\ref{M_EQ5}) contain fixed linear terms:
\begin{equation}
\label{M_EQ15}
\begin{split}
\frac{\text{d} \widehat{\mathcal{A}}(t)}{\text{d} t}  = - \epsilon \widehat{\mathcal{A}}(t) + (\cdot).
\end{split}
\end{equation}
These terms can, in essence, be computed exactly. However, they contain a series of exponential terms of the form $e^{-\epsilon \cdot (\cdot)}$. These exponential terms are fully preserved in Eqs.~(\ref{M_EQ13}, \ref{M_EQ14}) and therefore do not require approximation. In contrast, in direct numerical methods for ODEs, such as Eq.~(\ref{M_EQ8}), these inherently exact exponential terms still need to be approximated. Therefore, we adopt the iterative numerical computation process in Eqs.~(\ref{M_EQ13}, \ref{M_EQ14}) as the fundamental solution scheme of our method.

For the numerical computation of the integrals in Eqs.~(\ref{M_EQ13}, \ref{M_EQ14}), we employ the trapezoidal rule for the sake of simplicity:
\begin{equation}
\label{M_EQ16}
\begin{split}
\int_{t_n}^{t_{n+1}} e^{-\epsilon(t_{n+1}-s)} q^a(s)\text{d}s \approx \\ \frac{t_{n+1} - t_n}{2} (q^a(t_{n+1}) + e^{-\epsilon(t_{n+1} - t_{n})} q^a(t_{n})),
\end{split}
\end{equation}

\begin{equation}
\label{M_EQ17}
\begin{split}
\int_{t_n}^{t_{n+1}} e^{-\epsilon(t_{n+1}-s)} q^p(s)\text{d}s \approx \\ \frac{t_{n+1} - t_n}{2} (q^p(t_{n+1}) + e^{-\epsilon(t_{n+1} - t_{n})} q^p(t_{n})). 
\end{split}
\end{equation}

It should be noted that, although many high-accuracy numerical integration methods are currently available, Eqs.~(\ref{M_EQ13}, \ref{M_EQ14}) have already reduced this numerical approximation process to the sufficiently small interval $[t_n, t_{n+1}]$, which is enough to ensure the accuracy of the numerical computation. Under this circumstance, usability and simplicity become more important; therefore, the trapezoidal rule is suitable.

\subsubsection{Transformation over the First Time Interval}
\label{SEC_TFTI}
Within the first time interval $t_0 \leq t \leq t_1$, we assume that:
\begin{equation}
\label{M_EQ18}
\begin{cases}
f^{ds}(t_0) =f^{ds},\\
\mathcal{A}(t_0) e^{\mathcal{P}(t_0)} = \text{FFT}(f^{ds}(t_0)),  
\end{cases}
\end{equation}
where $f^{ds}$ is the domain-specific features obtained by backbone network as mentioned before. 

Furthermore, since this is the first time interval, especially at the left endpoint $t_0$, we do not have any prior information. Therefore, we make the following assumption: 
\begin{equation}
\label{M_EQ19}
\begin{cases}
\widehat{\mathcal{A}}(t_0) = q^a(t_0) = \mathcal{A}(t_0),\\
\widehat{\mathcal{P}}(t_0) = q^p(t_0) = \mathcal{P}(t_0).  
\end{cases}
\end{equation}
In this way, we define an initial value problem for Eq.~(\ref{M_EQ5}), and also specify $q^a(t_0)$ and $q^p(t_0)$ at the left endpoint $t_0$ in Eqs.~(\ref{M_EQ16}) and (\ref{M_EQ17}). For $q^a(t_1)$ and $q^p(t_1)$ at the right endpoint $t_1$, according to the definitions in Eqs.~(\ref{M_EQ3}) and (\ref{M_EQ4}), we define $\ddot{\mathcal{A}}(\ddot{t}_1)$ and $\ddot{\mathcal{P}}(\ddot{t}_1)$ as follows: 
\begin{equation}
\label{M_EQ20}
\begin{cases}
{\ddot{\mathcal{A}}(\ddot{t}_1)[c] = \mathcal{A}}^r(t_0)[c] = \frac{{\mathcal{A}}(t_0)[c]-\mu_c^{a}}{\sigma_c^a} \sigma_c^{a, r} + \mu_c^{a, r},\\
{\ddot{\mathcal{P}}(\ddot{t}_1)[c] = \mathcal{P}}(t_0)[c].
\end{cases}
\end{equation}
where $[c]$ gets the $c^{th}$ channel of a spectrum. $\mu_c^a$ and $\sigma_c^a$ are the mean and standard variance of ${\mathcal{A}}(t_0)[c]$. $\mu_c^{a, r}$ and $\sigma_c^{a, r}$ are obtained as follows: 
\begin{equation}
\label{M_EQ21}
\begin{cases}
\mu_c^{a, r} = (\alpha + 0.5) \cdot \mu_c^a,\\
\sigma_c^{a, r} = (1.5 - \alpha) \cdot \sigma_c^a, 
\end{cases}
\end{equation}
Here, $0\leq \alpha \leq 1$ is a randomly selected perturbation factor. Consequently, $q^a(t_1)$ and $q^p(t_1)$ are determined.

By combining Eqs.~(\ref{M_EQ13}, \ref{M_EQ14}) with Eqs.~(\ref{M_EQ16}, \ref{M_EQ17}), we obtain the estimation of the domain-agnostic spectra over the first time interval: 
\begin{equation}
\label{M_EQ22}
\begin{split}
\widehat{\mathcal{A}}(t_{1}) \approx e^{-\epsilon(t_{1}-t_0)} \widehat{\mathcal{A}}(t_0) + \\ \frac{\epsilon(t_{1} - t_0)}{2} (q^a(t_{1}) + e^{-\epsilon(t_1 - t_0)}q^a(t_{0})), 
\end{split}
\end{equation}
\begin{equation}
\label{M_EQ23}
\begin{split}
\widehat{\mathcal{P}}(t_{1}) \approx e^{-\epsilon(t_{1}-t_0)} \widehat{\mathcal{P}}(t_0) + \\ \frac{\epsilon(t_{1} - t_0)}{2} (q^p(t_{1}) + e^{-\epsilon(t_1 - t_0)}q^p(t_{0})).
\end{split}
\end{equation}

\subsubsection{Transformation over the Subsequent Time Intervals}
\label{SEC_TSTI}

The transformation over each subsequent time interval $[t_n, t_{n+1}]$ strictly follows Eqs.~(\ref{M_EQ13}-\ref{M_EQ14}), and the overall procedure is similar to that over the first time interval $[t_0, t_1]$. 
\begin{equation}
\label{M_EQ24}
\begin{split}
\widehat{\mathcal{A}}(t_{n+1}) \approx e^{-\epsilon(t_{n+1}-t_n)} \widehat{\mathcal{A}}(t_n) + \\ \frac{\epsilon(t_{n+1} - t_n)}{2} (q^a(t_{n+1}) + e^{-\epsilon(t_{n+1} - t_n)}q^a(t_{n})), 
\end{split}
\end{equation}
\begin{equation}
\label{M_EQ25}
\begin{split}
\widehat{\mathcal{P}}(t_{n+1}) \approx e^{-\epsilon(t_{n+1}-t_n)} \widehat{\mathcal{P}}(t_n) + \\ \frac{\epsilon(t_{n+1} - t_n)}{2} (q^p(t_{n+1}) + e^{-\epsilon(t_{n+1} - t_n)}q^p(t_{n})).
\end{split}
\end{equation}

Meanwhile, there are two main differences. First, we no longer need to compute the initial values of the domain-agnostic spectra. Instead, we only need to use $\widehat{\mathcal{A}}(t_n)$ and $\widehat{\mathcal{P}}(t_n)$ obtained from the previous time interval. Second, the computation of $q^a(t)$ and $q^p(t)$, where $t_n \leq t \leq t_{n+1}$, is different. Here, in conjunction with Eqs.~(\ref{M_EQ3}) and (\ref{M_EQ4}), and based on the knowledge accumulated from the transformation process over the preceding small time intervals, we make the following settings: 
\begin{equation}
\label{M_EQ26}
\begin{cases}
\ddot{\mathcal{A}}(\ddot{t}_{n}) = \widehat{\mathcal{A}}^r(t_{n-1}),\\
\ddot{\mathcal{P}}(\ddot{t}_{n}) = \widehat{\mathcal{P}}(t_{n-1}),\\
\ddot{\mathcal{A}}(\ddot{t}_{n+1}) = \widehat{\mathcal{A}}^r(t_{n}),\\
\ddot{\mathcal{P}}(\ddot{t}_{n+1}) = \widehat{\mathcal{P}}(t_{n}), 
\end{cases}
\end{equation}
where $\cdot^r$ is the feature perturbation operator as shown in Eq.~(\ref{M_EQ20}), In this way, $q^a(t_n)$, $q^p(t_n)$, $q^a(t_{n+1})$ and $q^p(t_{n+1})$ are determined based on Eqs.~(\ref{M_EQ3}) and (\ref{M_EQ4}). Consequently, the domain-agnostic spectra $\widehat{\mathcal{A}}(t_{n+1})$ and $\widehat{\mathcal{P}}(t_{n+1})$ are estimated according to Eqs. (\ref{M_EQ24}) and (\ref{M_EQ25}). 

When all the time intervals in the sequence $\{t_1, \cdots, t_n, \cdots, T\}$ are processed, the final approximations of domain-agnostic spectra $\widehat{\mathcal{A}}(T)$ and $\widehat{\mathcal{P}}(T)$ are computed. After IFFT is performed as Eq. (\ref{M_EQ2}), the domain-agnostic feature $f^{da}$ is obtained.

\subsection{Segmentation based on the Domain-Agnostic Features}
\label{SEC_SDAF}
In this section, we present the segmentation process in the source domain in detail. The segmentation process in the target domain is similar to that in the source domain. After the transformation over the time intervals $\{t_0, \cdots, t_i, \cdots, T\}$, the domain-agnostic features $f_k^{s, s, da}$ and $f^{s, q, da}$ for support and query images $I_k^{s, s, c^s}$ and $I^{s, q, c^s}$ in the source domain are obtained, where $k\in \{1, \cdots, K\}$. We obtain the self-support prototype and the adaptive self-support background prototype \cite{FSS-SSP-2022} and then engage in prototype and pixel matching to meet the objective of FSS. First, the support prototypes $\zeta_k^{s, s, f, c^s}$ and $\zeta_k^{s, s, b, c^s}$ for the foreground and background are generated using the masked average pooling (MAP) operation: 
\begin{equation}
\label{M_EQ27}
\begin{cases}
\zeta_k^{s, s, f, c^s} = \text{MAP}(M_k^{s, s, c^s}, f_k^{s, s, da}),\\
\zeta_k^{s, s, b, c^s} = \text{MAP}(1-M_k^{s, s, c^s}, f_k^{s, s, da}),
\end{cases}
\end{equation}
where $M_k^{s, s, c^s}$ is the mask of the support image for base category $c^s$. The prediction of $I^{s, q, c^s}$ is obtained based on the cosine similarity. 
\begin{equation}
\label{M_EQ28}
\begin{cases}
P_k^{s, q, f, c^s} = \text{cosine}(\zeta_k^{s, s, f, c^s}, f^{s, q, da}),\\
P_k^{s, q, b, c^s} = \text{cosine}(\zeta_k^{s, s, b, c^s}, f^{s, q, da}),
\end{cases}
\end{equation}
For each position $(i, j)$ in the image plane, the mask $\widehat{M}_k^{s, q, c^s}$ for the prediction is determined as follows:
\begin{equation}
\label{M_EQ29}
\widehat{M}_k^{s, q, c^s}(i, j) = 
\begin{cases}
1, & \text{if}\ P_k^{s, q, f, c^s}(i, j) > P_k^{s, q, b, c^s}(i, j),\\
0, & \text{otherwise}. 
\end{cases}
\end{equation}
Next, the self-support foreground prototype $\zeta_k^{s, q, f, c^s}$ is generated based on $\widehat{M}_k^{s, q, c^s}$ and the domain-agnostic feature $f^{s, q, da}$ of the query image $I^{s, q, c^s}$. 
\begin{equation}
\label{M_EQ30}
\zeta_k^{s, q, f, c^s} = \text{MAP}(\widehat{M}_k^{s, q, c^s}, f^{s, q, da}),
\end{equation}
To generate the adaptive self-support background prototype $\zeta_k^{s, q, b, c^s}$, the background query feature $f_k^{s, q, b, c^s}\in \mathbb{R}^{C\times N^{s, q, b, c^s}}$ is gathered: 
\begin{equation}
\label{M_EQ31}
f_k^{s, q, b, c^s} = (1 - \widehat{M}_k^{s, q, c^s}) \ast f^{s, q, da},
\end{equation}
where $\ast$ is the element-by-element multiplication. Then, $\zeta_k^{s, q, b, c^s}$ is obtained as follows:
\begin{equation}
\label{M_EQ32}
\zeta_k^{s, q, b, c^s} = f_k^{s, q, b, c^s} \text{softmax}({f_k^{s, q, b, c^s}}^T f^{s, q, da}),
\end{equation}

If it is a $K$-shot task, the support prototypes $\zeta^{s, s, f, c^s}$ and $\zeta^{s, s, b, c^s}$, the self-support foreground prototype $\zeta^{s, q, f, c^s}$, and the adaptive self-support background prototype $\zeta^{s, q, b, c^s}$ are computed as follows:
\begin{equation}
\label{M_EQ33}
\begin{cases}
\zeta^{s, s, f, c^s} = \frac{1}{K}\sum_{k=1}^K \zeta_k^{s, s, f, c^s},\\
\zeta^{s, s, b, c^s} = \frac{1}{K}\sum_{k=1}^K \zeta_k^{s, s, b, c^s},\\
\zeta^{s, q, f, c^s} = \frac{1}{K}\sum_{k=1}^K \zeta_k^{s, q, f, c^s},\\
\zeta^{s, q, b, c^s} = \frac{1}{K}\sum_{k=1}^K \zeta_k^{s, q, b, c^s}.
\end{cases}
\end{equation}

Then, $\zeta^{s, s, f, c^s}$, $\zeta^{s, s, b, c^s}$, $\zeta^{s, q, f, c^s}$, and $\zeta^{s, q, b, c^s}$ are combined:
\begin{equation}
\label{M_EQ34}
\begin{cases}
\zeta^{s, f, c^s} = \alpha_1 \zeta^{s, s, f, c^s} + \alpha_2 \zeta^{s, q, f, c^s},\\
\zeta^{s, b, c^s} = \alpha_1 \zeta^{s, s, b, c^s} + \alpha_2 \zeta^{s, q, b, c^s},
\end{cases}
\end{equation}
where $\alpha_1$ and $\alpha_2$ are tuning weights set according to \cite{FSS-SSP-2022}. 
Based on $\zeta^{s, f, c^s}$ and $\zeta^{s, b, c^s}$, the segmentation result for the query image $I^{s, q, c^s}$ on category $c^s$ is obtained using a similar method as Eqs. (\ref{M_EQ28}-\ref{M_EQ29}). 

In the target domain, we still provide a support set $S^{c^t}=\{I_k^{t, s, c^t}, M_k^{t, s, c^t}\}_{k=1}^K$ for each novel category $c^t$ for $K$-shot tasks when segmenting query images $\{I_i^{t, q}, i=1, \cdots, N^{t, q}\}$. We obtain the combined prototypes $\zeta^{t, f, c^t}$ and $\zeta^{t, b, c^t}$ after TTIs over $\{t_0, \cdots, t_i, \cdots, T\}$. Then, the segmentation result of each query image on the novel category $c^t$ is obtained in a similar way as Eqs. (\ref{M_EQ28}-\ref{M_EQ29}).

\subsection{Optimization in the Source Domain}
\label{SEC_REG2OPT}
The optimization process of the model ensures both the expressive power of domain-specific and domain-agnostic features in the source domain. To ensure the expressive power of domain-specific features, we use the domain-specific features of the support set $[{f_k^{s, s, ds}, k=1, \cdots, K}]$ to generate the support prototypes $\zeta^{s, s, f, c^s}$ and $\zeta^{s, s, b, c^s}$. 
\begin{equation}
\label{M_EQ35}
\begin{cases}
\zeta^{s, s, f, c^s} = \frac{1}{K}\sum_{k=1}^K \zeta_k^{s, s, f, c^s},\\
\zeta^{s, s, b, c^s} = \frac{1}{K}\sum_{k=1}^K \zeta_k^{s, s, b, c^s}.
\end{cases}
\end{equation}
where $\zeta_k^{s, s, f, c^s}$ and $\zeta_k^{s, s, b, c^s}$ are generated as (\ref{M_EQ27}). Then, the foreground and background prediction results of the query image $I^{s, q, c^s}$ are generated using Equation (\ref{M_EQ28}) and concatenated as $P^{s, q, ds, c^s}$. Subsequently, the loss $\mathcal{L}^{s, q, ds, c^s}$ of the domain-specific features is computed for backpropagation. 
\begin{equation}
\label{M_EQ36}
\mathcal{L}^{s, q, ds, c^s} = \text{BCE}(P^{s, q, ds, c^s}, M^{s, q, c^s}), 
\end{equation}
where $M^{s, q, c^s}$ is the ground-truth of $I^{s, q, c^s}$ for the category $c^s$. $\text{BCE}(\cdot, \cdot)$ computes the binary cross-entropy loss. 

To enforce the robustness of domain-agnostic features, we use the support prototypes, self-support foreground prototypes, and the adaptive self-support background prototypes, which are detailed in Sec.\ref{SEC_SDAF} to produce the prediction results $P^{s, q, da, c^s}$ and $\widehat{P}^{s, q, da, c^s}$ of the query image $I^{s, q, c^s}$ individually. Furthermore, we produce the prediction results $\widehat{P}_k^{s, s, da, c^s}$, $k = 1, \cdots, K$ for the support images $I_k^{s, s, c^s}$, $k = 1, \cdots, K$. 
Then, the losses $\mathcal{L}^{s, q, da, c^s}$ and $\mathcal{L}^{s, s, da, c^s}$ for the domain-agnostic features are computed: 
\begin{equation}
\label{M_EQ37}
\begin{split}
\mathcal{L}^{s, q, da, c^s} = \text{BCE}(P^{s, q, da, c^s}, M^{s, q, c^s}) \\ + \text{BCE}(\widehat{P}^{s, q, da, c^s}, M^{s, q, c^s}), 
\end{split}
\end{equation}
\begin{equation}
\label{M_EQ38}
\mathcal{L}^{s, s, da, c^s} = \sum_{k=1}^K \text{BCE}(P_k^{s, s, da, c^s}, M_k^{s, s, c^s}), 
\end{equation}
where $M_k^{s, s, c^s}, k=1, \cdots, K$ are the ground-truths of the support images $I_k^{s, s, c^s}, k = 1, \cdots, K$.  

In each training loop, a randomly selected category $c^s$ in the source domain is processed, and the following total loss is backpropagated to optimize the model. 
\begin{equation}
\label{M_EQ39}
\mathcal{L}^{c^s} = \mathcal{L}^{s, q, ds, c^s} + \mathcal{L}^{s, q, da, c^s} + \mathcal{L}^{s, s, da, c^s}. 
\end{equation}
After all the training loops are finished, all the categories in the source domain are processed by a large range of times, and the parameters in the model are optimized.

\begin{table*}[h]
	\footnotesize
	\centering
	\caption[]{Employed datasets}
	\begin{tabular}{m{2cm}m{15cm}}
	\toprule
	\makecell*[c]{Datasets} & \makecell*[c]{Descriptions} \\ 
	\midrule
	\makecell*[c]{PASCAL VOC} & A natural image dataset containing 20 categories and a background with Semantic Boundary Dataset (SBD) \cite{BSD-2011} augmentation. \\ 
	\makecell*[c]{DeepGlobe} & A satellite RS dataset containing 6 categories and an unknown background, with a spatial resolution of 0.5m, mainly covering rural areas. The introduction of this dataset helps analyze the reliability of the model in satellite RS applications \cite{OTHER-Hong2024}. \\ 
	\makecell*[c]{ISIC} & A medical image dataset containing three types of skin lesions, used to evaluate the ability of few-shot segmentation methods to identify cancer on the human skin surface. \\ 
	\makecell*[c]{Chest X-Ray} & A medical image dataset for collecting tuberculosis. Images are collected from 58 abnormal cases exhibiting signs of tuberculosis and 80 normal cases. \\ 
	\makecell*[c]{FSS-1000} & A natural image dataset specialized for FSS tasks. FSS-1000 is considerably challenged because it contains 1000 categories, and scarce or tiny objects.  \\
	\midrule
	\makecell*[c]{Potsdam} & An aerial RS image dataset containing 5 categories and a background, with a spatial resolution of approximately 0.5 m, covering the Potsdam city, in which extremely small ground targets are present.  \\
	\makecell*[c]{FBP} & A dataset constructed from Gaofen-2 (GF-2) satellite imagery, with a spatial resolution of 4m, covering more than 50,000 square kilometers across China and containing 24 fine-grained land-cover categories. \\
	\makecell*[c]{LoveDA} & A high-resolution satellite imagery dataset with a spatial resolution of 0.3m, covering urban and rural areas, and containing 6 categories and a background. We combine all the images in both urban and rural areas for test. \\
	\makecell*[c]{iSAID} & A RS dataset collected from multiple aerial and satellite platforms with spatial resolutions from 0.1m to 4m, containing 15 categories covering typical man-made objects.  \\
	\bottomrule
	\end{tabular}
	\label{TAB_DATA}
\end{table*}

\begin{table*}[h]
	\centering
	\caption[]{The cross-domain tasks with different domain shifts.}
	\begin{tabular}{*{5}l}
		\toprule
		& \multirow{2}{*}{\makecell[l]{Cross-domain tasks\\ (source $\rightarrow$ target)}} & \multicolumn{3}{c}{Domain shifts} \\
		\cline{3-5}
		& & \makecell[c]{Scene organization} & \makecell[c]{Target characteristics} & \makecell[c]{Spatial resolution} \\
		\midrule
		\multirow{10}{*}{\makecell*[l]{Target domains\\ of standard \\CD-FSS datasets:}} & PASCAL VOC $\rightarrow$ DeepGlobe & \makecell[c]{human-view $\rightarrow$ satellite-view} & \makecell[c]{daily-life targets $\rightarrow$ land cover} & \makecell[c]{high $\rightarrow$ low} \\
		& PASCAL VOC $\rightarrow$ ISIC & \makecell[c]{human-view $\rightarrow$ skin-surface \\imaging} & \makecell[c]{daily-life targets $\rightarrow$ skin cancer} & \makecell[c]{low $\rightarrow$ high} \\
		& PASCAL VOC $\rightarrow$ Chest X-Ray & \makecell[c]{human-view $\rightarrow$ X-Ray imaging} & \makecell[c]{daily-life targets $\rightarrow$ chest} & \makecell[c]{low $\rightarrow$ high} \\
		& PASCAL VOC $\rightarrow$ FSS-1000 & \makecell[c]{--} & \makecell[c]{few targets $\rightarrow$ many targets} & \makecell[c]{--} \\
		\cline{2-5}
		& DeepGlobe $\rightarrow$ PASCAL VOC & \makecell[c]{satellite-view $\rightarrow$ human-view} & \makecell[c]{land cover $\rightarrow$ daily-life targets} & \makecell[c]{low $\rightarrow$ high} \\
		& DeepGlobe $\rightarrow$ ISIC & \makecell[c]{satellite-view $\rightarrow$ skin-surface \\imaging} & \makecell[c]{land cover $\rightarrow$ skin cancer} & \makecell[c]{low $\rightarrow$ high} \\
		& DeepGlobe $\rightarrow$ Chest X-Ray & \makecell[c]{satellite-view $\rightarrow$ X-Ray imaging} & \makecell[c]{land cover $\rightarrow$ chest} & \makecell[c]{low $\rightarrow$ high} \\
		& DeepGlobe $\rightarrow$ FSS-1000 & \makecell[c]{satellite-view $\rightarrow$ human-view} & \makecell[c]{land cover $\rightarrow$ daily-life targets} & \makecell[c]{low $\rightarrow$ high} \\
		\midrule
		\multirow{13}{*}{\makecell*[l]{Target domains \\of RS datasets:}} & PASCAL VOC $\rightarrow$ Potsdam & \makecell[c]{human-view $\rightarrow$ aerial view\\ daily-life $\rightarrow$ land-surface scenes} & \makecell[c]{daily-life targets $\rightarrow$ land cover \\ \& small targets} & \makecell[c]{high $\rightarrow$ low} \\
		& PASCAL VOC $\rightarrow$ FBP & \makecell[c]{human-view $\rightarrow$ satellite-view\\ daily-life $\rightarrow$ land-surface scenes} & \makecell[c]{daily-life targets $\rightarrow$ \\fine-grained land cover} & \makecell[c]{high $\rightarrow$ low} \\
		& PASCAL VOC $\rightarrow$ LoveDA & \makecell[c]{human-view $\rightarrow$ satellite-view\\ daily-life $\rightarrow$ land-surface scenes} & \makecell[c]{daily-life targets $\rightarrow$ coarse land cover} & \makecell[c]{high $\rightarrow$ low} \\
		& PASCAL VOC $\rightarrow$ iSAID & \makecell[c]{human-view $\rightarrow$ multi-views\\ daily-life $\rightarrow$ land-surface scenes} & \makecell[c]{daily-life targets $\rightarrow$ man-made\\ targets} & \makecell[c]{high $\rightarrow$ low} \\
		\cline{2-5}
		& DeepGlobe $\rightarrow$ Potsdam & \makecell[c]{satellite-view $\rightarrow$ aerial view} & \makecell[c]{land cover $\rightarrow$\\ multi-scale targets} & \makecell[c]{--}\\
		& DeepGlobe $\rightarrow$ FBP & \makecell[c]{--} & \makecell[c]{coarse categories $\rightarrow$\\ fine-grained categories} & \makecell[c]{high $\rightarrow$ low} \\
		& DeepGlobe $\rightarrow$ LoveDA & \makecell[c]{rural $\rightarrow$ rural \& urban} & \makecell[c]{land cover category systems\\ differ to some extent} & \makecell[c]{low $\rightarrow$ high} \\
		& DeepGlobe $\rightarrow$ iSAID & \makecell[c]{satellite-view $\rightarrow$ diverse views\\ rural $\rightarrow$ urban} & \makecell[c]{land cover $\rightarrow$ man-made targets} & \makecell[c]{$0.5m$ $\rightarrow$\\ $[0.1m, 4.0m]$} \\
		\bottomrule
	\end{tabular}
	\label{TAB_TASKS}
\end{table*}

\begin{table*}[h]
	\footnotesize
	\centering
	\caption[]{Comparison methods.}
	\begin{tabular}{m{2cm}m{0.8cm}m{1cm}m{8cm}m{4cm}}
		\toprule
		\makecell*[c]{Methods} & \makecell*[l]{Para.} & \makecell*[l]{GFLOPs} & \makecell*[c]{Modules} & \makecell*[c]{Optimization Process} \\
		\midrule
		\makecell*[c]{PATNet \cite{CDFSS-Lei2022}} & 28.14 & 153.87 & (1) Pyramid anchor-based transformation module (PATM), (2) domain-agnostic correlation learning, (3) 4D convolutional
pyramid encoder, and (4) 2D convolutional context decoder. & Source-domain training $+$ Target-domain fine-tuning (S.$+$T.). \\ 
		\makecell*[c]{PMNet \cite{CDFSS-Chen2024a}} & 37.25 & 132.10 & (1) Hysteretic spatial filtering module and (2) bidirectional 3D convolution-based pixel-to-pixel and pixel-to-patch relationship capture. & Source-domain training only (S.O.). \\ 
		\makecell*[c]{APM-M \cite{CDFSS-Tong2024}} & 28.13 & 119.07 & Amplitude-phase masker (APM) module and (2) adaptive channel phase attention (ACPA) module. & Source-domain training $+$ Target-domain fine-tuning (S.$+$T.). \\ 
		\makecell*[c]{ABCDFSS \cite{CDFSS-Herzog2024}} & 25.56 & 795.58 & (1) Adapter network (which is appended to the backbone), (2) self-supervised embedding alignment, (3) class alignment, and (4) dense comparison. & Target-domain adaptation only (T.O.). \\ 
		\makecell*[c]{DR-Adapter \cite{CDFSS-Su2024}} & 59.25 & 513.29 & (1) Local domain perturbation, (2) global domain perturbation, (3) domain rectification, and (4) cyclic domain alignment. & Source-domain training only (S.O.). \\ 
		\makecell*[c]{DMTNet \cite{CDFSS-Chen2024b}} & 28.15 & 238.17 & (1) Similarity-based self-matching, (2) adaptive feature transformation, and (3) dual hypercorrelation construction. & Source-domain training $+$ Target-domain fine-tuning (S.$+$T.). \\ 
		\makecell*[c]{IFA \cite{CDFSS-Nie2024}} & 8.67 & 133.62 & (1) Bi-directional few-shot prediction (BFP) and (2) iterative few-shot adaptor. & Source-domain training $+$ Target-domain fine-tuning (S.$+$T.). \\
		\midrule
		\makecell*[c]{FSS-TIs} & 19.90 & 83.44 & TTIs. & Source-domain training $+$ Target-domain fine-tuning (S.$+$T.). \\ 
		\bottomrule
	\end{tabular}
	\label{TAB_CM}
\end{table*}

\subsection{Fine-Tuning and Testing in the Target Domain}
As described in Sec. \ref{SEC_MO}, for the $K$-shot task, the target-domain fine-tuning and testing processes provide only $K$ samples for each novel category as support information. These constitute all the information available for training the model during fine-tuning. Furthermore, the query images and their corresponding masks used for loss computation are also randomly and automatically generated from these $K$ support samples. The specific procedure of target-domain fine-tuning is illustrated in Fig. \ref{FIG_OVERVIEW}. To avoid redundancy, we do not repeat the details here. 

In the testing process, for the $K$-shot task, only the same set of support information used in the target domain fine-tuning process is employed (with only $K$ support samples for each novel category). This ensures consistency with the requirements of real-world tasks and avoids the need for additional manual annotations. Unlike the source-domain training processes, TTIs does not add random spectral perturbations during testing and target-domain fine-tuning.

\section{Experiments}
\label{SEC_EXPERIMENTS}

\subsection{Datasets and Cross-domain Tasks}
\label{SEC_DATASET}

The experimental data are divided into two parts. The first part consists of five mainstream datasets currently used to evaluate CD-FSS methods: PASCAL VOC \cite{PASCAL-2010, PASCAL-2015}, DeepGlobe \cite{DeepGlobe-2018}, ISIC \cite{ISIC-IEEE-2018, ISIC-SD-2018, ISIC-ARXIV-2019}, Chest X-ray \cite{Chest-TMI-2014a, Chest-TMI-2014b}, and FSS-1000 \cite{FSS-2020}. The second part consists of four remote sensing (RS) datasets from distinctly different scenarios: Potsdam \cite{Potsdam-Rottensteiner2014}, FBP \cite{FBP-Tong2023}, LoveDA \cite{LoveDA-Wang2021}, and iSAID \cite{iSAID-Zamir2019}. Detailed information of these datasets is shown in Table \ref{TAB_DATA}. From Table \ref{TAB_DATA}, we can find that these datasets are from definitely different domains.

To comprehensively evaluate the robustness of the proposed method against different types of domain shifts, we construct 16 distinctive cross-domain tasks based on the above nine datasets (as shown in Table~\ref{TAB_TASKS}). Among them, PASCAL VOC $\rightarrow$ DeepGlobe, PASCAL VOC $\rightarrow$ ISIC, PASCAL VOC $\rightarrow$ Chest X-Ray, and PASCAL VOC $\rightarrow$ FSS-1000 are currently the standard tasks for evaluating CD-FSS models. However, as can be observed from Table~\ref{TAB_TASKS}, these standard tasks all adopt a single source-domain setting, and the source-domain distribution may overlap with that of the target domain. Therefore, based on these five standard datasets, we further design four additional cross-domain tasks, namely, DeepGlobe $\rightarrow$ PASCAL VOC, DeepGlobe $\rightarrow$ ISIC, DeepGlobe $\rightarrow$ Chest X-Ray, and DeepGlobe $\rightarrow$ FSS-1000. In these four tasks, the source and target domains differ significantly in three aspects, namely scene organization, target characteristics, and spatial resolution. Furthermore, to evaluate stability under extreme domain shifts, we additionally introduce four RS datasets from different scenarios to construct the tasks PASCAL VOC $\rightarrow$ Potsdam, PASCAL VOC $\rightarrow$ FBP, PASCAL VOC $\rightarrow$ LoveDA, and PASCAL VOC $\rightarrow$ iSAID, thereby transferring the model from the scale of daily human life to a more macroscopic scale. Finally, DeepGlobe is used as the source domain and the four RS datasets are considered as target domains. As DeepGlobe is also a RS dataset, the domain shifts in these four cross-domain tasks are not as severe as those when PASCAL VOC is used as the source domain. Therefore, the comparison of the accuracies of the cross-domain tasks from PASCAL VOC and DeepGlobe to the four RS datasets can further verify the influence of the magnitude of domain shifts between the source and target domains on FSS-TIs.

\subsection{Evaluation metrics}
\label{SEC_METRICS}
To evaluate the performance of FSS-TIs and quantitatively compare it with existing methods, we follow the studies in \cite{CDFSS-Lei2022, CDFSS-Chen2024b, CDFSS-Nie2024}, which employ the mean intersection over union (mIoU) metric. Furthermore, because the $K$ support samples of each novel category are randomly selected in the target-domain dataset when we fine tune and test the model, each method is tested 20 times. Then, the mean value and standard deviation of all the mIoU scores for these tests are determined for comparison. The higher the mean value of the mIoU scores and the smaller the standard deviation, the better the performance of the method.

\subsection{Implementation Details}
\label{SEC_IMPLEMENTATION}

Since FSS-TIs involves FFT and the iterative numerical computation process of ODEs, an increase in computational burden is unavoidable. To address this issue, before applying FFT, we compress the channel number of the domain-specific features to half of the original channel number, and then generate domain-agnostic features for few-shot segmentation. Regarding the time intervals $\{t_0, \cdots, t_i, \cdots, T\}$, we set $T=t_{5}$ and use an equal interval setting. The default interval is 0.01, and the initial value of $\epsilon$ in Eq. (\ref{M_EQ5}) is set to $0.5$ by default. Here, $T = t_{5}$ means that TTIs performs iterative transformations over 5 intervals. Experimental results show that as the number of intervals increases, the accuracy of FSS-TIs improves. Once it reaches $T = t_{5}$, the accuracy does not improve much as the number of intervals increases. Therefore, we set $T = t_{5}$ as the default value.

\subsection{Comparison Methods}
\label{SEC_CM}

We compare our FSS-TIs with seven representative CD-FSS methods, including PATNet \cite{CDFSS-Lei2022}, PMNet \cite{CDFSS-Chen2024a}, APM-M \cite{CDFSS-Tong2024}, ABCDFSS \cite{CDFSS-Herzog2024}, DR-Adapter \cite{CDFSS-Su2024}, DMTNet \cite{CDFSS-Chen2024b}, and IFA \cite{CDFSS-Nie2024} to validate the effectiveness of our study. Detailed information about these CD-FSS methods can be found in Table \ref{TAB_CM}. As shown in Table \ref{TAB_CM}, FSS-TIs achieves the most concise design of all methods by proposing only all-in-one module TTIs, ensuring simplicity and ease of use. 

In addition, we calculate the number of parameters (Para., the unit is million (M)) and the computational complexity (measured in GFLOPs) of all methods. In computing these parameter counts and computational complexities, ResNet50 is used as the backbone for all methods (since all models adopt ResNet50 as the only default backbone or as one of their default backbones). The input data size is kept consistent, that is, the dimensions of both the support and query images are set to $[1, 3, 400, 400]$, while the dimensions of the label maps are set to $[1, 400, 400]$, and the 1-shot setting is used. It can be observed that the parameter count of FSS-TIs is only larger than that of IFA and is clearly smaller than those of the other existing methods, while its GFLOPs score is the lowest. Notably, although the numerical computation process of FSS-TIs includes iterative steps, which intuitively would increase the computational complexity, our experiments show that only five iterations are sufficient to achieve accuracy superior to that of existing methods. In addition, a range of existing methods modify certain convolutional layers inside ResNet50 to increase the feature resolution, which leads to higher computational complexity. By contrast, FSS-TIs does not make any modification to the ResNet50 architecture provided by PyTorch, but only uses its first three stages for feature extraction. Combined with the concise structure of FSS-TIs and the aforementioned channel compression (as shown in Sec.~\ref{SEC_IMPLEMENTATION}), this ultimately results in lower GFLOPs score.

\subsection{Results}
\label{SEC_COMPARISON}

\subsubsection{Comparison on Standard CD-FSS Datasets}
\label{SEC_COMPARISON_STANDARD}
In this part, we provide the comparison on two sets of cross-domain tasks (as shown in the upper half of Table \ref{TAB_TASKS}). As is well known, PASCAL VOC is a natural image dataset, while DeepGlobe is a satellite RS image dataset. There are significant differences between the two in scene organization, target characteristics, and spatial resolution. Using each of them as the source domain to define the two sets of cross-domain tasks can more comprehensively reflect the performance and stability of different methods.. 

Tables \ref{TAB_QC-SET1} and \ref{TAB_QC-SET2} show the quantitative comparisons between our FSS-ITs and existing CD-FSS methods. Some existing methods use not only ResNet50 as the backbone but also VGG-16. Therefore, we tested the accuracy of these methods when using ResNet50 and VGG-16 separately. For fairness of comparison, our FSS-ITs was also tested with both backbones. From the quantitative comparison results in tables \ref{TAB_QC-SET1} and \ref{TAB_QC-SET2}, it can be observed that whether PASCAL VOC or DeepGlobe is used as the source domain, FSS-ITs consistently achieves higher accuracy than existing methods. More specifically, for all methods that use both ResNet50 and VGG-16 as backbones, accuracy is higher when using ResNet50, and FSS-ITs is no exception. Moreover, when comparing the accuracy values between tables \ref{TAB_QC-SET1} and \ref{TAB_QC-SET2}, we find that almost all methods achieve slightly higher testing accuracy on the ISIC2018, Chest X-Ray, and FSS-1000 datasets when trained with PASCAL VOC as the source domain. However, the difference is insignificant. This indicates that although the domain shift between satellite RS images in DeepGlobe and the images in the other datasets is more severe, the cross-domain segmentation capability of the CD-FSS method is not significantly affected through source-domain training and target-domain fine-tuning. In addition, in terms of stability—that is, from the comparison of mIoU scores' standard deviation (``Dev." in the tables)—the choice between ResNet50 and VGG-16 has little impact on the stability of the methods. The choice of the source-domain dataset also has little impact. However, when comparing the standard deviations across methods, we can still conclude that FSS-ITs exhibits higher stability. 

\begin{table*}[h]
	\scriptsize
	\centering
	\caption[]{Quantitative results on standard CD-FSS datasets based on the mIoU (\%) metric using PASCAL VOC as the source domain.}
	\begin{tabular}{*{18}l}
		\toprule
		\multicolumn{18}{c}{Source Domain: PASCAL VOC $\rightarrow$ Target Domain: Below} \\
		\midrule
		\multirow{3}{*}{Methods} & \multirow{3}{*}{Backbone} & \multicolumn{4}{c}{DeepGlobe} & \multicolumn{4}{c}{ISIC} & \multicolumn{4}{c}{Chest X-Ray} & \multicolumn{4}{c}{FSS-1000} \\
		\cline{3-18}
		& & \multicolumn{2}{c}{$1$-shot} & \multicolumn{2}{c}{$5$-shot} & \multicolumn{2}{c}{$1$-shot} & \multicolumn{2}{c}{$5$-shot} & \multicolumn{2}{c}{$1$-shot} & \multicolumn{2}{c}{$5$-shot} & \multicolumn{2}{c}{$1$-shot} & \multicolumn{2}{c}{$5$-shot}  \\
		\cline{3-18}
		& & \makecell*[c]{Mean} & \makecell*[c]{Dev.} & \makecell*[c]{Mean} & \makecell*[c]{Dev.} & \makecell*[c]{Mean} & \makecell*[c]{Dev.} & \makecell*[c]{Mean} & \makecell*[c]{Dev.} & \makecell*[c]{Mean} & \makecell*[c]{Dev.} & \makecell*[c]{Mean} & \makecell*[c]{Dev.} & \makecell*[c]{Mean} & \makecell*[c]{Dev.} & \makecell*[c]{Mean} & \makecell*[c]{Dev.}\\
		\midrule
		\makecell*[l]{PATNet \cite{CDFSS-Lei2022}} & \makecell*[c]{Vgg-16} & \makecell*[c]{28.55} & \makecell*[c]{4.49} & \makecell*[c]{34.13} & \makecell*[c]{4.32} & \makecell*[c]{31.93} & \makecell*[c]{4.21} & \makecell*[c]{41.09} & \makecell*[c]{4.02} & \makecell*[c]{55.19} & \makecell*[c]{2.94} & \makecell*[c]{58.22} & \makecell*[c]{2.85} & \makecell*[c]{69.32} & \makecell*[c]{0.97} & \makecell*[c]{73.44} & \makecell*[c]{0.85} \\ 
		\makecell*[l]{DMTNet \cite{CDFSS-Chen2024b}} & \makecell*[c]{Vgg-16} & \makecell*[c]{33.89} & \makecell*[c]{4.72} & \makecell*[c]{45.53} & \makecell*[c]{4.46} & \makecell*[c]{33.27} & \makecell*[c]{4.14} & \makecell*[c]{40.04} & \makecell*[c]{4.11} & \makecell*[c]{65.26} & \makecell*[c]{3.43} & \makecell*[c]{70.05} & \makecell*[c]{3.39} & \makecell*[c]{69.66} & \makecell*[c]{0.95} & \makecell*[c]{72.47} & \makecell*[c]{0.93} \\ 
		\makecell*[l]{FSS-TIs} & \makecell*[c]{Vgg-16} & \makecell*[c]{\bf 38.76} & \makecell*[c]{\bf 3.47} & \makecell*[c]{\bf 46.41} & \makecell*[c]{\bf 3.68} & \makecell*[c]{\bf 38.69} & \makecell*[c]{\bf 3.73} & \makecell*[c]{\bf 42.14} & \makecell*[c]{\bf 3.75} & \makecell*[c]{\bf 67.16} & \makecell*[c]{\bf 2.82} & \makecell*[c]{\bf 71.54} & \makecell*[c]{\bf 2.68} & \makecell*[c]{\bf 70.16} & \makecell*[c]{\bf 0.86} & \makecell*[c]{\bf 74.13} & \makecell*[c]{\bf 0.74} \\ 
		\midrule
		\makecell*[l]{PATNet \cite{CDFSS-Lei2022}} & \makecell*[c]{ResNet50} & \makecell*[c]{37.89} & \makecell*[c]{4.12} & \makecell*[c]{42.97} & \makecell*[c]{3.92} & \makecell*[c]{41.16} & \makecell*[c]{3.95} & \makecell*[c]{53.58} & \makecell*[c]{3.74} & \makecell*[c]{66.61} & \makecell*[c]{2.88} & \makecell*[c]{70.20} & \makecell*[c]{2.63} & \makecell*[c]{75.56} & \makecell*[c]{0.93} & \makecell*[c]{78.43} & \makecell*[c]{0.89}  \\ 
		\makecell*[l]{PMNet \cite{CDFSS-Chen2024a}} & \makecell*[c]{ResNet50} & \makecell*[c]{37.10} & \makecell*[c]{4.45} & \makecell*[c]{41.60} & \makecell*[c]{4.23} & \makecell*[c]{51.20} & \makecell*[c]{3.84} & \makecell*[c]{54.5} & \makecell*[c]{3.32} & \makecell*[c]{70.4} & \makecell*[c]{2.94} & \makecell*[c]{74.0} & \makecell*[c]{2.67} & \makecell*[c]{77.33} & \makecell*[c]{0.94} & \makecell*[c]{81.23} & \makecell*[c]{0.67} \\ 
		\makecell*[l]{APM-M \cite{CDFSS-Tong2024}} & \makecell*[c]{ResNet50} & \makecell*[c]{40.08} & \makecell*[c]{4.32} & \makecell*[c]{43.89} & \makecell*[c]{4.02} & \makecell*[c]{41.07} & \makecell*[c]{3.51} & \makecell*[c]{50.68} & \makecell*[c]{3.22} & \makecell*[c]{75.71} & \makecell*[c]{2.49} & \makecell*[c]{79.02} & \makecell*[c]{2.88} & \makecell*[c]{74.29} & \makecell*[c]{0.93} & \makecell*[c]{76.27} & \makecell*[c]{0.82}\\ 
		\makecell*[l]{ABCDFSS \cite{CDFSS-Herzog2024}} & \makecell*[c]{ResNet50} & \makecell*[c]{41.68} & \makecell*[c]{8.69} & \makecell*[c]{47.09} & \makecell*[c]{8.20} & \makecell*[c]{44.57} & \makecell*[c]{8.64} & \makecell*[c]{52.39} & \makecell*[c]{8.27} & \makecell*[c]{76.73} & \makecell*[c]{7.92} & \makecell*[c]{79.49} & \makecell*[c]{7.66} & \makecell*[c]{74.60} & \makecell*[c]{4.84} & \makecell*[c]{76.20} & \makecell*[c]{4.79}\\ 
		\makecell*[l]{DR-Adapter \cite{CDFSS-Su2024}} & \makecell*[c]{ResNet50} & \makecell*[c]{40.62} & \makecell*[c]{4.53} & \makecell*[c]{47.46} & \makecell*[c]{4.39} & \makecell*[c]{40.13} & \makecell*[c]{3.89} & \makecell*[c]{46.86} & \makecell*[c]{3.87} & \makecell*[c]{77.45} & \makecell*[c]{2.85} & \makecell*[c]{79.49} & \makecell*[c]{2.47} & \makecell*[c]{75.73} & \makecell*[c]{0.98} & \makecell*[c]{78.42} & \makecell*[c]{0.96}\\ 
		\makecell*[l]{DMTNet \cite{CDFSS-Chen2024b}} & \makecell*[c]{ResNet50} & \makecell*[c]{42.11} & \makecell*[c]{4.22} & \makecell*[c]{49.33} & \makecell*[c]{4.13} & \makecell*[c]{46.66} & \makecell*[c]{3.75} & \makecell*[c]{54.01} & \makecell*[c]{3.24} & \makecell*[c]{76.52} & \makecell*[c]{2.17} & \makecell*[c]{78.39} & \makecell*[c]{2.12} & \makecell*[c]{76.53} & \makecell*[c]{1.01} & \makecell*[c]{79.87} & \makecell*[c]{0.85} \\ 
		\makecell*[l]{IFA \cite{CDFSS-Nie2024}} & \makecell*[c]{ResNet50} & \makecell*[c]{43.96} & \makecell*[c]{4.79} & \makecell*[c]{51.73} & \makecell*[c]{4.16} & \makecell*[c]{48.66} & \makecell*[c]{3.97} & \makecell*[c]{54.21} & \makecell*[c]{3.93} & \makecell*[c]{76.33} & \makecell*[c]{2.87} & \makecell*[c]{78.61} & \makecell*[c]{2.75} & \makecell*[c]{77.64} & \makecell*[c]{1.05} & \makecell*[c]{80.31} & \makecell*[c]{0.79}\\ 
		\makecell*[l]{FSS-TIs (ours)} & \makecell*[c]{ResNet50} & \makecell*[c]{\bf 46.52} & \makecell*[c]{\bf 3.88} & \makecell*[c]{\bf 54.73} & \makecell*[c]{\bf 3.22} & \makecell*[c]{\bf 51.94} & \makecell*[c]{\bf 3.49} & \makecell*[c]{\bf 55.26} & \makecell*[c]{3.29} & \makecell*[c]{\bf 78.66} & \makecell*[c]{\bf 2.14} & \makecell*[c]{\bf 81.23} & \makecell*[c]{\bf 2.08} & \makecell*[c]{\bf 79.37} & \makecell*[c]{\bf 0.84} & \makecell*[c]{\bf 82.60} & \makecell*[c]{\bf 0.61} \\ 
		\bottomrule
	\end{tabular}
	\label{TAB_QC-SET1}
\end{table*}

\begin{table*}[h]
	\scriptsize
	\centering
	\caption[]{Quantitative results on standard CD-FSS datasets based on the mIoU (\%) metric using DeepGlobe as the source domain.}
	\begin{tabular}{*{18}l}
		\toprule
		\multicolumn{18}{c}{Source Domain: DeepGlobe $\rightarrow$ Target Domain: Below} \\
		\midrule
		\multirow{3}{*}{Methods} & \multirow{3}{*}{Backbone} & \multicolumn{4}{c}{PASCAL VOC} & \multicolumn{4}{c}{ISIC} & \multicolumn{4}{c}{Chest X-Ray} & \multicolumn{4}{c}{FSS-1000} \\
		\cline{3-18}
		& & \multicolumn{2}{c}{$1$-shot} & \multicolumn{2}{c}{$5$-shot} & \multicolumn{2}{c}{$1$-shot} & \multicolumn{2}{c}{$5$-shot} & \multicolumn{2}{c}{$1$-shot} & \multicolumn{2}{c}{$5$-shot} & \multicolumn{2}{c}{$1$-shot} & \multicolumn{2}{c}{$5$-shot}  \\
		\cline{3-18}
		& & \makecell*[c]{Mean} & \makecell*[c]{Dev.} & \makecell*[c]{Mean} & \makecell*[c]{Dev.} & \makecell*[c]{Mean} & \makecell*[c]{Dev.} & \makecell*[c]{Mean} & \makecell*[c]{Dev.} & \makecell*[c]{Mean} & \makecell*[c]{Dev.} & \makecell*[c]{Mean} & \makecell*[c]{Dev.} & \makecell*[c]{Mean} & \makecell*[c]{Dev.} & \makecell*[c]{Mean} & \makecell*[c]{Dev.}\\
		\midrule
		\makecell*[l]{PATNet \cite{CDFSS-Lei2022}} & \makecell*[c]{Vgg-16} & \makecell*[c]{40.56} & \makecell*[c]{4.49} & \makecell*[c]{47.17} & \makecell*[c]{4.37} & \makecell*[c]{30.94} & \makecell*[c]{4.19} & \makecell*[c]{40.73} & \makecell*[c]{4.07} & \makecell*[c]{54.67} & \makecell*[c]{3.16} & \makecell*[c]{57.71} & \makecell*[c]{2.99} & \makecell*[c]{68.86} & \makecell*[c]{1.31} & \makecell*[c]{73.02} & \makecell*[c]{1.18} \\ 
		\makecell*[l]{DMTNet \cite{CDFSS-Chen2024b}} & \makecell*[c]{Vgg-16} & \makecell*[c]{41.87} & \makecell*[c]{4.33} & \makecell*[c]{49.28} & \makecell*[c]{4.19} & \makecell*[c]{31.24} & \makecell*[c]{4.26} & \makecell*[c]{41.39} & \makecell*[c]{4.03} & \makecell*[c]{64.61} & \makecell*[c]{2.85} & \makecell*[c]{69.96} & \makecell*[c]{2.79} & \makecell*[c]{69.29} & \makecell*[c]{1.04} & \makecell*[c]{72.21} & \makecell*[c]{0.96} \\ 
		\makecell*[l]{FSS-TIs} & \makecell*[c]{Vgg-16} & \makecell*[c]{\bf 45.19} & \makecell*[c]{\bf 3.92} & \makecell*[c]{\bf 50.27} & \makecell*[c]{\bf 3.83} & \makecell*[c]{\bf 39.03} & \makecell*[c]{\bf 3.91} & \makecell*[c]{\bf 42.07} & \makecell*[c]{\bf 3.68} & \makecell*[c]{\bf 66.93} & \makecell*[c]{\bf 2.77} & \makecell*[c]{\bf 70.99} & \makecell*[c]{\bf 2.64} & \makecell*[c]{\bf 69.78} & \makecell*[c]{\bf 0.94} & \makecell*[c]{\bf 73.86} & \makecell*[c]{\bf 0.83} \\ 
		\midrule
		\makecell*[l]{PATNet \cite{CDFSS-Lei2022}} & \makecell*[c]{ResNet50} & \makecell*[c]{48.35} & \makecell*[c]{4.22} & \makecell*[c]{51.25} & \makecell*[c]{4.02} & \makecell*[c]{39.38} & \makecell*[c]{3.94} & \makecell*[c]{48.47} & \makecell*[c]{3.62} & \makecell*[c]{64.89} & \makecell*[c]{2.79} & \makecell*[c]{68.34} & \makecell*[c]{2.14} & \makecell*[c]{76.02} & \makecell*[c]{1.46} & \makecell*[c]{79.37} & \makecell*[c]{1.32}  \\ 
		\makecell*[l]{PMNet \cite{CDFSS-Chen2024a}} & \makecell*[c]{ResNet50} & \makecell*[c]{48.69} & \makecell*[c]{4.36} & \makecell*[c]{51.92} & \makecell*[c]{4.21} & \makecell*[c]{46.39} & \makecell*[c]{3.89} & \makecell*[c]{50.67} & \makecell*[c]{3.74} & \makecell*[c]{65.77} & \makecell*[c]{2.43} & \makecell*[c]{69.48} & \makecell*[c]{2.18} & \makecell*[c]{76.94} & \makecell*[c]{1.27} & \makecell*[c]{79.04} & \makecell*[c]{1.18} \\ 
		\makecell*[l]{APM-M \cite{CDFSS-Tong2024}} & \makecell*[c]{ResNet50} & \makecell*[c]{49.57} & \makecell*[c]{4.62} & \makecell*[c]{53.67} & \makecell*[c]{4.16} & \makecell*[c]{39.97} & \makecell*[c]{3.73} & \makecell*[c]{45.38} & \makecell*[c]{3.44} & \makecell*[c]{75.20} & \makecell*[c]{2.76} & \makecell*[c]{78.44} & \makecell*[c]{2.79} & \makecell*[c]{74.03} & \makecell*[c]{1.23} & \makecell*[c]{76.11} & \makecell*[c]{1.08}\\ 
		\makecell*[l]{ABCDFSS \cite{CDFSS-Herzog2024}} & \makecell*[c]{ResNet50} & \makecell*[c]{50.16} & \makecell*[c]{8.43} & \makecell*[c]{53.99} & \makecell*[c]{8.26} & \makecell*[c]{44.57} & \makecell*[c]{8.64} & \makecell*[c]{52.39} & \makecell*[c]{8.27} & \makecell*[c]{76.73} & \makecell*[c]{7.92} & \makecell*[c]{79.49} & \makecell*[c]{7.66} & \makecell*[c]{74.60} & \makecell*[c]{4.84} & \makecell*[c]{76.20} & \makecell*[c]{4.79}\\ 
		\makecell*[l]{DR-Adapter \cite{CDFSS-Su2024}} & \makecell*[c]{ResNet50} & \makecell*[c]{49.33} & \makecell*[c]{4.17} & \makecell*[c]{52.67} & \makecell*[c]{4.02} & \makecell*[c]{38.83} & \makecell*[c]{3.84} & \makecell*[c]{45.92} & \makecell*[c]{3.63} & \makecell*[c]{77.25} & \makecell*[c]{2.93} & \makecell*[c]{79.52} & \makecell*[c]{2.82} & \makecell*[c]{74.63} & \makecell*[c]{1.25} & \makecell*[c]{78.38} & \makecell*[c]{1.17}\\ 
		\makecell*[l]{DMTNet \cite{CDFSS-Chen2024b}} & \makecell*[c]{ResNet50} & \makecell*[c]{52.14} & \makecell*[c]{4.09} & \makecell*[c]{55.88} & \makecell*[c]{3.94} & \makecell*[c]{46.45} & \makecell*[c]{3.97} & \makecell*[c]{53.33} & \makecell*[c]{3.78} & \makecell*[c]{75.52} & \makecell*[c]{2.66} & \makecell*[c]{77.01} & \makecell*[c]{2.26} & \makecell*[c]{77.56} & \makecell*[c]{1.18} & \makecell*[c]{80.36} & \makecell*[c]{0.91} \\ 
		\makecell*[l]{IFA \cite{CDFSS-Nie2024}} & \makecell*[c]{ResNet50} & \makecell*[c]{53.59} & \makecell*[c]{4.52} & \makecell*[c]{57.73} & \makecell*[c]{4.16} & \makecell*[c]{48.37} & \makecell*[c]{3.92} & \makecell*[c]{54.06} & \makecell*[c]{3.88} & \makecell*[c]{76.11} & \makecell*[c]{2.76} & \makecell*[c]{77.92} & \makecell*[c]{2.59} & \makecell*[c]{77.51} & \makecell*[c]{1.24} & \makecell*[c]{80.25} & \makecell*[c]{0.96}\\ 
		\makecell*[l]{FSS-TIs (Ours)} & \makecell*[c]{ResNet50} & \makecell*[c]{\bf 55.08} & \makecell*[c]{\bf 3.75} & \makecell*[c]{\bf 59.17} & \makecell*[c]{\bf 3.29} & \makecell*[c]{\bf 51.82} & \makecell*[c]{\bf 3.52} & \makecell*[c]{\bf 55.13} & \makecell*[c]{\bf 3.36} & \makecell*[c]{\bf 78.61} & \makecell*[c]{\bf 2.17} & \makecell*[c]{\bf 81.07} & \makecell*[c]{\bf 2.13} & \makecell*[c]{\bf 79.19} & \makecell*[c]{\bf 0.96} & \makecell*[c]{\bf 82.61} & \makecell*[c]{\bf 0.74} \\ 
		\bottomrule
	\end{tabular}
	\label{TAB_QC-SET2}
\end{table*}

\begin{figure}[!t]
\centering
\includegraphics[width=3.5in]{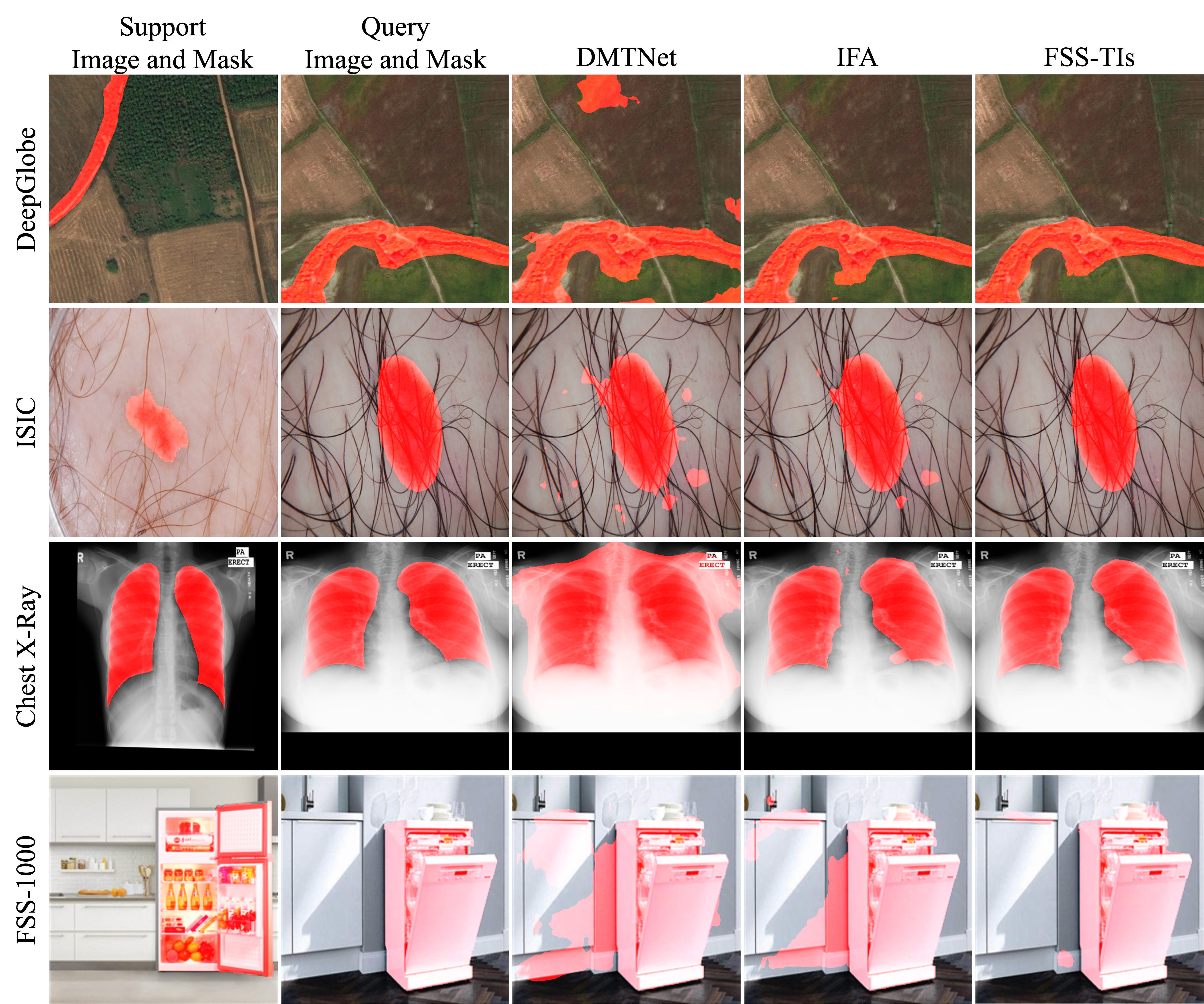}
\caption{The visual comparisons between FSS-TIs and existing methods on the first set of CD-FSS tasks which uses PASCAL VOC as the source-domain dataset.}
\label{FIG_VC-P}
\end{figure}

\begin{figure}[!t]
\centering
\includegraphics[width=3.5in]{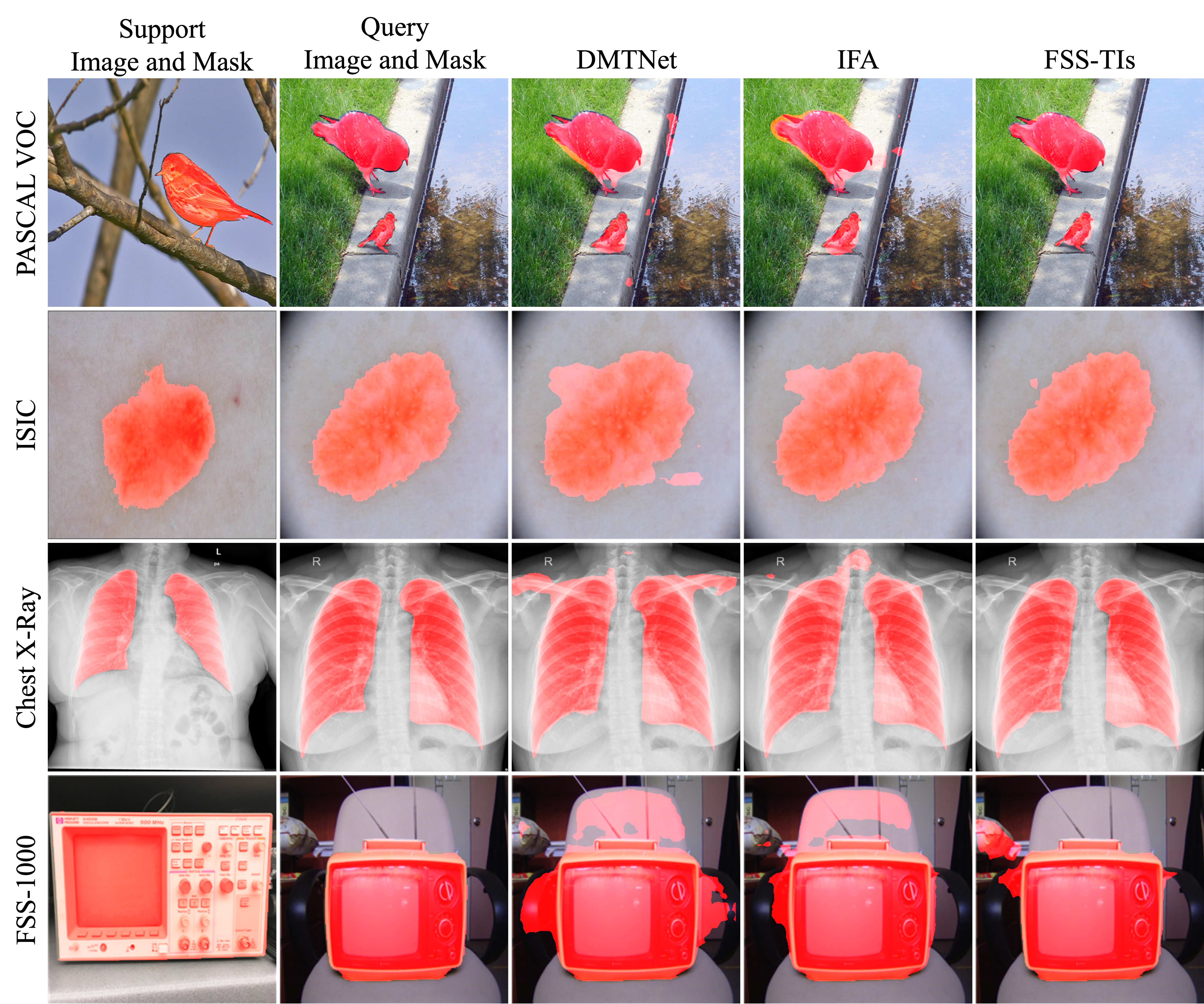}
\caption{The visual comparisons between FSS-TIs and existing methods on the second set of CD-FSS tasks which uses DeepGlobe as the source-domain dataset.}
\label{FIG_VC-D}
\end{figure}

From the quantitative comparison results in tables \ref{TAB_QC-SET1} and \ref{TAB_QC-SET2}, it can be witnessed that DMTNet and IFA, both using ResNet50 as the backbone, achieve the highest accuracy among existing methods. Therefore, in this section, we mainly compare the visual results of our method with those of DMTNet and IFA. Figs. \ref{FIG_VC-P} and \ref{FIG_VC-D} present the visualization comparison results. Due to space limitations, only the visualization results for the $1$-shot task are conveyed here. In each figure, the first column represents the support information, specifically including one support image and its corresponding mask. The second column depicts the query image and its corresponding mask. The last three columns display the segmentation results of DMTNet, IFA, and FSS-TIs, respectively. Since each dataset contains multitudinous images and it is difficult to present all of them here, the segmentation results of only one representative query image are shown for each dataset. From Fig. \ref{FIG_VC-P} and Fig. \ref{FIG_VC-D}, it can be observed that our method achieves superior visualized segmentation results, which corroborates the accuracy comparison results in Table \ref{TAB_QC-SET1} and Table \ref{TAB_QC-SET2}. 

\subsubsection{Comparison on More Diverse Datasets}
\label{SEC_COMPARISON_DIVERSE}

To evaluate the cross-domain performance of the proposed method under more extreme domain shifts, this paper further introduces diverse RS datasets to construct two groups of cross-domain tasks (as shown in the lower half of Table~\ref{TAB_TASKS}). In the first group of cross-domain tasks, PASCAL VOC is used as the source domain to evaluate the transfer performance from human daily-life scenes in natural images to diverse RS scenarios. In the second group of cross-domain tasks, DeepGlobe is used as the source domain, while the diverse RS datasets are respectively treated as the target domains. Since DeepGlobe itself is also a satellite RS dataset, the severity of domain shift in this group of cross-domain tasks is lower than that in the first group. The comparison between the two groups can therefore be used to analyze the impact of domain-shift severity.

Tables~\ref{TAB_QC-SET3} and \ref{TAB_QC-SET4} present the results and accuracy comparisons for the first and second groups of cross-domain tasks. As shown in Tables~\ref{TAB_QC-SET1} and \ref{TAB_QC-SET2}, the accuracy of all methods with ResNet50 as the backbone is higher than that with VGG-16 as the backbone. Therefore, only the results of all methods using ResNet50 as the backbone are reported here. From the comparison among different methods, FSS-TIs consistently achieves the best segmentation accuracy, and in most cases also yields the smallest standard deviation of segmentation accuracy. Furthermore, from the accuracy comparison between the two groups of cross-domain tasks, the difference between using PASCAL VOC as the source domain and using DeepGlobe as the source domain is not large. This indicates that the amount of domain shifts does not have a substantial impact on FSS-TIs. However, our intuition suggests that the more severe the domain shift is, the lower the segmentation accuracy will be. The result is inconsistent with our intuition. To this end, in the Ablation Study and Discussion (Sec.~\ref{SEC_ABS}), we provide a detailed explanation of why FSS-TIs is insensitive to the amount of domain shift.

\begin{table*}[h]
	\scriptsize
	\centering
	\caption[]{Quantitative results on diverse RS datasets based on the mIoU (\%) metric using PASCAL VOC as the source domain.}
	\begin{tabular}{*{17}l}
		\toprule
		\multicolumn{17}{c}{Source Domain: PASCAL VOC $\rightarrow$ Target Domain: Below} \\
		\midrule
		\multirow{3}{*}{Methods} & \multicolumn{4}{c}{Potsdam} & \multicolumn{4}{c}{FBP} & \multicolumn{4}{c}{LoveDA} & \multicolumn{4}{c}{iSAID} \\
		\cline{2-17}
		& \multicolumn{2}{c}{$1$-shot} & \multicolumn{2}{c}{$5$-shot} & \multicolumn{2}{c}{$1$-shot} & \multicolumn{2}{c}{$5$-shot} & \multicolumn{2}{c}{$1$-shot} & \multicolumn{2}{c}{$5$-shot} & \multicolumn{2}{c}{$1$-shot} & \multicolumn{2}{c}{$5$-shot}  \\
		\cline{2-17}
		& \makecell*[c]{Mean} & \makecell*[c]{Dev.} & \makecell*[c]{Mean} & \makecell*[c]{Dev.} & \makecell*[c]{Mean} & \makecell*[c]{Dev.} & \makecell*[c]{Mean} & \makecell*[c]{Dev.} & \makecell*[c]{Mean} & \makecell*[c]{Dev.} & \makecell*[c]{Mean} & \makecell*[c]{Dev.} & \makecell*[c]{Mean} & \makecell*[c]{Dev.} & \makecell*[c]{Mean} & \makecell*[c]{Dev.}\\
		\midrule
		\makecell*[l]{PATNet \cite{CDFSS-Lei2022}} & \makecell*[c]{32.83} & \makecell*[c]{4.75} & \makecell*[c]{37.65} & \makecell*[c]{4.26} & \makecell*[c]{20.17} & \makecell*[c]{3.89} & \makecell*[c]{27.58} & \makecell*[c]{3.62} & \makecell*[c]{33.57} & \makecell*[c]{4.36} & \makecell*[c]{39.26} & \makecell*[c]{4.29} & \makecell*[c]{17.54} & \makecell*[c]{4.71} & \makecell*[c]{22.67} & \makecell*[c]{4.42}  \\ 
		\makecell*[l]{PMNet \cite{CDFSS-Chen2024a}} & \makecell*[c]{33.52} & \makecell*[c]{4.67} & \makecell*[c]{38.28} & \makecell*[c]{4.16} & \makecell*[c]{25.49} & \makecell*[c]{3.83} & \makecell*[c]{31.17} & \makecell*[c]{3.44} & \makecell*[c]{36.19} & \makecell*[c]{4.51} & \makecell*[c]{43.41} & \makecell*[c]{4.07} & \makecell*[c]{20.17} & \makecell*[c]{4.62} & \makecell*[c]{26.42} & \makecell*[c]{4.38} \\ 
		\makecell*[l]{APM-M \cite{CDFSS-Tong2024}} & \makecell*[c]{35.96} & \makecell*[c]{4.54} & \makecell*[c]{40.62} & \makecell*[c]{4.23} & \makecell*[c]{27.75} & \makecell*[c]{3.67} & \makecell*[c]{34.06} & \makecell*[c]{3.56} & \makecell*[c]{37.53} & \makecell*[c]{4.31} & \makecell*[c]{44.83} & \makecell*[c]{3.96} & \makecell*[c]{20.67} & \makecell*[c]{4.33} & \makecell*[c]{26.76} & \makecell*[c]{4.17}\\ 
		\makecell*[l]{ABCDFSS \cite{CDFSS-Herzog2024}} & \makecell*[c]{36.12} & \makecell*[c]{6.94} & \makecell*[c]{41.04} & \makecell*[c]{6.23} & \makecell*[c]{27.28} & \makecell*[c]{5.44} & \makecell*[c]{33.11} & \makecell*[c]{5.31} & \makecell*[c]{37.13} & \makecell*[c]{7.95} & \makecell*[c]{44.26} & \makecell*[c]{7.62} & \makecell*[c]{21.32} & \makecell*[c]{8.62} & \makecell*[c]{27.36} & \makecell*[c]{8.37}\\ 
		\makecell*[l]{DR-Adapter \cite{CDFSS-Su2024}} & \makecell*[c]{33.72} & \makecell*[c]{4.36} & \makecell*[c]{38.37} & \makecell*[c]{4.09} & \makecell*[c]{26.85} & \makecell*[c]{3.89} & \makecell*[c]{32.91} & \makecell*[c]{3.53} & \makecell*[c]{36.16} & \makecell*[c]{4.25} & \makecell*[c]{43.45} & \makecell*[c]{3.71} & \makecell*[c]{20.82} & \makecell*[c]{4.42} & \makecell*[c]{26.43} & \makecell*[c]{4.22}\\ 
		\makecell*[l]{DMTNet \cite{CDFSS-Chen2024b}} & \makecell*[c]{36.01} & \makecell*[c]{4.29} & \makecell*[c]{40.82} & \makecell*[c]{4.13} & \makecell*[c]{28.67} & \makecell*[c]{3.53} & \makecell*[c]{34.29} & \makecell*[c]{3.11} & \makecell*[c]{38.72} & \makecell*[c]{4.31} & \makecell*[c]{45.62} & \makecell*[c]{3.85} & \makecell*[c]{22.16} & \makecell*[c]{4.38} & \makecell*[c]{28.13} & \makecell*[c]{4.13} \\ 
		\makecell*[l]{IFA \cite{CDFSS-Nie2024}} & \makecell*[c]{36.25} & \makecell*[c]{4.57} & \makecell*[c]{40.68} & \makecell*[c]{4.36} & \makecell*[c]{29.17} & \makecell*[c]{3.42} & \makecell*[c]{34.93} & \makecell*[c]{3.16} & \makecell*[c]{39.26} & \makecell*[c]{4.12} & \makecell*[c]{46.13} & \makecell*[c]{3.94} & \makecell*[c]{21.67} & \makecell*[c]{4.66} & \makecell*[c]{27.83} & \makecell*[c]{4.27}\\ 
		\makecell*[l]{FSS-TIs (ours)} & \makecell*[c]{\bf 39.09} & \makecell*[c]{\bf 3.94} & \makecell*[c]{\bf 45.18} & \makecell*[c]{\bf 3.73} & \makecell*[c]{\bf 31.55} & \makecell*[c]{\bf 2.63} & \makecell*[c]{\bf 36.11} & \makecell*[c]{\bf 2.51} & \makecell*[c]{\bf 41.96} & \makecell*[c]{\bf 3.24} & \makecell*[c]{\bf 48.36} & \makecell*[c]{\bf 3.17} & \makecell*[c]{\bf 24.51} & \makecell*[c]{\bf 3.62} & \makecell*[c]{\bf 31.21} & \makecell*[c]{\bf 3.34} \\ 
		\bottomrule
	\end{tabular}
	\label{TAB_QC-SET3}
\end{table*}

\begin{table*}[h]
	\scriptsize
	\centering
	\caption[]{Quantitative results on diverse RS datasets based on the mIoU (\%) metric using DeepGlobe as the source domain.}
	\begin{tabular}{*{17}l}
		\toprule
		\multicolumn{17}{c}{Source Domain: DeepGlobe $\rightarrow$ Target Domain: Below} \\
		\midrule
		\multirow{3}{*}{Methods} & \multicolumn{4}{c}{Potsdam} & \multicolumn{4}{c}{FBP} & \multicolumn{4}{c}{LoveDA} & \multicolumn{4}{c}{iSAID} \\
		\cline{2-17}
		& \multicolumn{2}{c}{$1$-shot} & \multicolumn{2}{c}{$5$-shot} & \multicolumn{2}{c}{$1$-shot} & \multicolumn{2}{c}{$5$-shot} & \multicolumn{2}{c}{$1$-shot} & \multicolumn{2}{c}{$5$-shot} & \multicolumn{2}{c}{$1$-shot} & \multicolumn{2}{c}{$5$-shot}  \\
		\cline{2-17}
		& \makecell*[c]{Mean} & \makecell*[c]{Dev.} & \makecell*[c]{Mean} & \makecell*[c]{Dev.} & \makecell*[c]{Mean} & \makecell*[c]{Dev.} & \makecell*[c]{Mean} & \makecell*[c]{Dev.} & \makecell*[c]{Mean} & \makecell*[c]{Dev.} & \makecell*[c]{Mean} & \makecell*[c]{Dev.} & \makecell*[c]{Mean} & \makecell*[c]{Dev.} & \makecell*[c]{Mean} & \makecell*[c]{Dev.}\\
		\midrule
		\makecell*[l]{PATNet \cite{CDFSS-Lei2022}} & \makecell*[c]{32.97} & \makecell*[c]{4.56} & \makecell*[c]{37.84} & \makecell*[c]{4.21} & \makecell*[c]{20.41} & \makecell*[c]{3.58} & \makecell*[c]{27.37} & \makecell*[c]{3.52} & \makecell*[c]{33.91} & \makecell*[c]{4.49} & \makecell*[c]{39.67} & \makecell*[c]{4.38} & \makecell*[c]{17.56} & \makecell*[c]{4.84} & \makecell*[c]{22.96} & \makecell*[c]{4.57}  \\ 
		\makecell*[l]{PMNet \cite{CDFSS-Chen2024a}} & \makecell*[c]{33.77} & \makecell*[c]{4.42} & \makecell*[c]{37.86} & \makecell*[c]{4.09} & \makecell*[c]{25.52} & \makecell*[c]{3.76} & \makecell*[c]{31.69} & \makecell*[c]{3.36} & \makecell*[c]{36.52} & \makecell*[c]{4.82} & \makecell*[c]{43.58} & \makecell*[c]{4.13} & \makecell*[c]{20.04} & \makecell*[c]{4.53} & \makecell*[c]{26.26} & \makecell*[c]{4.13} \\ 
		\makecell*[l]{APM-M \cite{CDFSS-Tong2024}} & \makecell*[c]{36.27} & \makecell*[c]{4.41} & \makecell*[c]{40.97} & \makecell*[c]{4.12} & \makecell*[c]{27.51} & \makecell*[c]{3.72} & \makecell*[c]{34.12} & \makecell*[c]{3.73} & \makecell*[c]{37.81} & \makecell*[c]{4.17} & \makecell*[c]{45.04} & \makecell*[c]{3.78} & \makecell*[c]{20.28} & \makecell*[c]{4.52} & \makecell*[c]{26.43} & \makecell*[c]{4.28}\\ 
		\makecell*[l]{ABCDFSS \cite{CDFSS-Herzog2024}} & \makecell*[c]{36.12} & \makecell*[c]{6.94} & \makecell*[c]{41.04} & \makecell*[c]{6.23} & \makecell*[c]{27.28} & \makecell*[c]{5.44} & \makecell*[c]{33.11} & \makecell*[c]{5.31} & \makecell*[c]{37.13} & \makecell*[c]{7.95} & \makecell*[c]{44.26} & \makecell*[c]{7.62} & \makecell*[c]{21.32} & \makecell*[c]{8.62} & \makecell*[c]{27.36} & \makecell*[c]{8.37}\\ 
		\makecell*[l]{DR-Adapter \cite{CDFSS-Su2024}} & \makecell*[c]{34.96} & \makecell*[c]{4.15} & \makecell*[c]{38.48} & \makecell*[c]{4.14} & \makecell*[c]{26.44} & \makecell*[c]{3.42} & \makecell*[c]{32.52} & \makecell*[c]{3.27} & \makecell*[c]{36.58} & \makecell*[c]{4.13} & \makecell*[c]{43.75} & \makecell*[c]{3.61} & \makecell*[c]{20.27} & \makecell*[c]{4.58} & \makecell*[c]{26.17} & \makecell*[c]{4.33}\\ 
		\makecell*[l]{DMTNet \cite{CDFSS-Chen2024b}} & \makecell*[c]{36.38} & \makecell*[c]{4.16} & \makecell*[c]{41.27} & \makecell*[c]{4.54} & \makecell*[c]{28.83} & \makecell*[c]{3.26} & \makecell*[c]{34.64} & \makecell*[c]{3.04} & \makecell*[c]{38.98} & \makecell*[c]{4.20} & \makecell*[c]{45.88} & \makecell*[c]{3.62} & \makecell*[c]{22.69} & \makecell*[c]{4.14} & \makecell*[c]{28.59} & \makecell*[c]{4.07} \\ 
		\makecell*[l]{IFA \cite{CDFSS-Nie2024}} & \makecell*[c]{36.63} & \makecell*[c]{4.43} & \makecell*[c]{40.93} & \makecell*[c]{4.52} & \makecell*[c]{28.39} & \makecell*[c]{3.35} & \makecell*[c]{35.17} & \makecell*[c]{3.11} & \makecell*[c]{39.77} & \makecell*[c]{4.01} & \makecell*[c]{46.38} & \makecell*[c]{3.72} & \makecell*[c]{21.91} & \makecell*[c]{4.47} & \makecell*[c]{28.03} & \makecell*[c]{4.15}\\ 
		\makecell*[l]{FSS-TIs (ours)} & \makecell*[c]{\bf 39.61} & \makecell*[c]{\bf 3.85} & \makecell*[c]{\bf 45.67} & \makecell*[c]{\bf 3.64} & \makecell*[c]{\bf 31.95} & \makecell*[c]{\bf 2.57} & \makecell*[c]{\bf 36.82} & \makecell*[c]{\bf 2.49} & \makecell*[c]{\bf 42.08} & \makecell*[c]{\bf 3.18} & \makecell*[c]{\bf 48.92} & \makecell*[c]{\bf 3.09} & \makecell*[c]{\bf 25.12} & \makecell*[c]{\bf 3.79} & \makecell*[c]{\bf 31.68} & \makecell*[c]{\bf 3.24} \\ 
		\bottomrule
	\end{tabular}
	\label{TAB_QC-SET4}
\end{table*}

\subsection{Ablation Study and Discussion}
\label{SEC_ABS}
\subsubsection{Component Analysis}
The all-in-one module named TTIs in FSS-ITs integrates the FFT, random spectral perturbation (RSP), and ODE into a feature transformation process over a sequence of time intervals. If TTIs is removed from the FSS-ITs, the cross-domain effect will also be erased. In our notation, the variant of our method without the ODE modeling is abbreviated FSS-TIs$-$ODE. The variant without FFT is abbreviated FSS-TIs$-$FFT, and the variant without RSP is abbreviated FSS-TIs$-$RSP. In addition, during the optimization process in Sec. \ref{SEC_REG2OPT}, we include a segmentation loss $\mathcal{L}^{s,q,ds,c^s}$ based on the raw features (domain-specific features) extracted by the backbone network. We also perform ablation studies on the loss. The variant without $\mathcal{L}^{s,q,ds,c^s}$ is denoted as FSS-TIs$-$$\mathcal{L}^{s,q,ds,c^s}$. The more severe the accuracy drops after removing a certain component, the more important that component is.

\begin{table}[h]
	\scriptsize
	\centering
	\caption[]{Ablation study on the components of FSS-TIs when DeepGlobe is the target-domain dataset and PASCAL VOC is the source-domain dataset.}
	\begin{tabular}{*{7}l}
	\toprule
	\multirow{2}{*}{Method} & \multicolumn{3}{c}{$1$-shot} & \multicolumn{3}{c}{$5$-shot} \\ 
	\cline{2-7}
	& \makecell*[c]{Mean} & \makecell*[c]{Dev.} & \makecell*[c]{FPS} & \makecell*[c]{Mean} & \makecell*[c]{Dev.} & \makecell*[c]{FPS} \\
	\midrule
	FSS-TIs$-$TTIs & 36.26 & 4.57 & 217.21 & 41.09 & 4.23 & 86.41 \\
	FSS-TIs$-$ODE  & 39.16 & 4.61 & 192.73 & 43.13 & 4.32  & 71.37 \\
	FSS-TIs$-$FFT  & 42.08 & 4.52 & 168.45 & 49.33 & 4.07  & 65.72 \\
	FSS-TIs$-$RSP  & 44.58 & 4.29 & 147.53 & 52.21 & 4.13  & 54.59 \\
	\makecell[l]{FSS-TIs$-$$\mathcal{L}^{s,q,ds,c^s}$}  & 45.63 & 3.96 & 147.53 & 54.13 & 3.62 & 54.59 \\
	FSS-TIs  & 46.52 & 3.88 & 147.53 & 54.73 & 3.31 & 54.59 \\
	\makecell[l]{FSS-TIs (WoST Phase)} & 43.11 & 6.24 & 148.61 & 50.76 & 5.45 & 55.27 \\
	\makecell[l]{FSS-TIs (explicit Euler)} & 42.33 & 4.55 & 147.96 & 50.12 & 3.96 & 54.88 \\
	\makecell[l]{FSS-TIs (improved Euler)} & 43.24 & 3.92 & 147.71 & 51.31 & 3.39 & 54.64 \\
	\bottomrule
	\end{tabular}
	\label{TAB_ABS-COMPONENTS}
\end{table}

Table \ref{TAB_ABS-COMPONENTS} presents the ablation study results for each component of FSS-TIs. From the accuracy comparison in Table \ref{TAB_ABS-COMPONENTS}, it can be seen that when the complete FSS-TIs is applied, the accuracy is the highest. When all four components are removed, the cross-domain settings would not be took into account, resulting in a large performance gap compared to FSS-TIs. By comparing the accuracies when only a single component is removed, the roles of ODE and FFT are highlighted, as ODE deepens the feature representation and FFT decomposes the features into amplitude and phase spectra so that they can be processed separately. RSP and $\mathcal{L}^{s,q,ds,c^s}$ also play positive roles to different extents. In summary, each component in FSS-TIs is necessary. In addition, we also evaluate a variant of FSS-TIs without special treatment of the phase, denoted as FSS-TIs (WoST Phase). In this variant, Eq.~(\ref{M_EQ4}) is not used to convert the phase into sine and cosine values. Instead, convolution is directly applied on phase spectrum to compute $q^p(t)$. We find that the accuracy of FSS-TIs (WoST Phase) drops, and the standard deviation of the accuracy also increases, which demonstrates the importance of Eq.~(\ref{M_EQ4}). To experimentally demonstrate the advantage of first deriving the analytical solution of the ODEs and then performing numerical computation, as in Eqs.~(\ref{M_EQ9}--\ref{M_EQ14}), we also implement two direct numerical computation processes for ODEs based on the explicit and improved Euler methods. These methods replace the corresponding ODE-related computation process in FSS-TIs, and the resulting accuracy is also evaluated, as shown by the last two rows in Table~\ref{TAB_ABS-COMPONENTS}. From the accuracy comparison, it can be observed that FSS-TIs (improved Euler) and FSS-TIs (explicit Euler) obtain lower accuracy than the default version of FSS-TIs adopted in this paper, which is consistent with the discussion in Section~\ref{SEC_NCODE}.

In addition, Table~\ref{TAB_ABS-COMPONENTS} also reports the FPS of different variants of FSS-TIs to measure the computational overhead introduced by each component. When computing FPS, the settings are consistent with that in Sec.~\ref{SEC_CM}. Since FPS is typically evaluated in inference mode, the computation of the loss function does not affect the inference speed. Meanwhile, the RSP is no longer introduced during inference, that is, RSP also does not affect the FPS values. Therefore, the FPS values in the $4^{th}$ to $6^{th}$ rows of Table~\ref{TAB_ABS-COMPONENTS} are identical. On the one hand, it can be observed from the FPS values that the inference speed of the complete FSS-TIs is sufficient. On the other hand, the ODE numerical computation and FFT in TTIs have a relatively large impact on FPS. However, it is precisely the ODE and FFT that play the leading role in improving accuracy. The reduction in FPS (i.e., the increase in computational overhead) caused by introducing ODE and FFT is exactly the cost of improving accuracy. We believe that this trade-off is worthwhile, because ODE and FFT provide theoretical interpretability and a concise network structure.

\begin{table*}[h]
	\scriptsize
	\centering
	\caption[]{Analysis of the sensitivity to the amount of domain shift and support sizes.}
	\begin{tabular}{*{12}l}
	\toprule
	\multirow{2}{*}{Tasks} & \multirow{2}{*}{Methods} & \multicolumn{2}{c}{$1$-shot} & \multicolumn{2}{c}{$3$-shot} & \multicolumn{2}{c}{$5$-shot} & \multicolumn{2}{c}{$8$-shot} & \multicolumn{2}{c}{$10$-shot} \\ 
	\cline{3-12}
	&  & \makecell*[c]{Mean} & \makecell*[c]{Dev.} & \makecell*[c]{Mean} & \makecell*[c]{Dev.} & \makecell*[c]{Mean} & \makecell*[c]{Dev.} & \makecell*[c]{Mean} & \makecell*[c]{Dev.} & \makecell*[c]{Mean} & \makecell*[c]{Dev.}\\
	\midrule
	\makecell*[l]{----\ $\rightarrow$ Potsdam} & \makecell*[l]{FSS-TIs$-$TTIs (T.O.)} & 33.35 & 9.83 & 35.62 & 9.76 & 37.95 & 9.74  & 39.79 & 9.54 & 40.23 & 9.27\\
	\makecell*[l]{----\ $\rightarrow$ Potsdam} & \makecell*[l]{FSS-TIs (T.O.)} & 37.91 & 8.67 & 40.83 & 8.29 & 43.74 & 8.27  & 47.35 & 8.30 & 49.91 & 8.11\\
	\multirow{4}{*}{\makecell*[l]{PASCAL VOC$\rightarrow$ Potsdam}} & \makecell*[l]{FSS-TIs$-$TTIs (S.O.)} & 29.58 & 4.82 & 31.71 & 4.93 & 33.24 & 4.66 & 34.59 & 4.48 & 35.07 & 4.29\\
	& \makecell*[l]{FSS-TIs (S.O.)} & 34.06 & 4.26 & 36.17 & 4.18 & 38.71 & 4.06 & 39.26 & 3.98 & 40.17 & 3.81\\
	& \makecell*[l]{FSS-TIs} & 39.09 & 3.94 & 41.92 & 4.03 & 45.18 & 3.73  & 48.17 & 3.59 & 50.96 & 3.56\\
	\multirow{4}{*}{\makecell*[l]{DeepGlobe $\rightarrow$ Potsdam}} & \makecell*[l]{FSS-TIs$-$TTIs (S.O.)} & 30.07 & 4.74 & 31.77 & 4.52 & 34.17 & 4.41 & 35.56 & 4.26 & 35.79 & 4.07\\
	& \makecell*[l]{FSS-TIs (S.O.)} & 36.11 & 4.06 & 37.81 & 3.98 & 39.85 & 3.67 & 40.62 & 3.71 & 40.93 & 3.72\\
	& \makecell*[l]{FSS-TIs} & 39.61 & 3.85  & 42.38 & 3.76 & 45.67 & 3.64  & 48.74 & 3.47 & 51.57 & 3.42\\
	\bottomrule
	\end{tabular}
	\label{TAB_ABS-SADS}
\end{table*}

\subsubsection{Why is FSS-TIs Insensitive to the Amount of Domain Shift?}

Sec.~\ref{SEC_COMPARISON_DIVERSE} has already led to the conclusion that FSS-TIs is insensitive to the magnitude of domain shift, which is inconsistent with our intuition. Here, a more detailed discussion of the experimental results on PASCAL VOC $\rightarrow$ Potsdam and DeepGlobe $\rightarrow$ Potsdam is provided to explain this phenomenon. PASCAL VOC and Potsdam belong to two completely different domains, and their only commonality is that both use images as the medium to present their respective scenes. In contrast, although DeepGlobe and Potsdam belong to different domains, they share many similarities. First, both are RS datasets and present scenes of the Earth's surface from a top-down view. Second, their category systems also exhibit certain similarities, such as agricultural land in DeepGlobe and low vegetation in Potsdam. Therefore, the domain shift faced by PASCAL VOC $\rightarrow$ Potsdam is clearly more severe, whereas the domain shift faced by DeepGlobe $\rightarrow$ Potsdam is milder. The comparison between these two tasks is therefore representative for analyzing the issue discussed in this section. 

As shown in Table~\ref{TAB_ABS-SADS}, we report the accuracy of the model trained only on the source domain without target-domain fine-tuning (FSS-TIs (S.O.)), as well as the accuracy of the model fine-tuned only on the target domain without source-domain training (FSS-TIs (T.O.)). We observe that, in the DeepGlobe $\rightarrow$ Potsdam task, the accuracy of FSS-TIs (S.O.) differs from that in the PASCAL VOC $\rightarrow$ Potsdam task. This gap is larger than that of FSS-TIs between the two tasks, especially under the 1-shot setting. This indicates that the severity of domain shift has an impact on the performance of FSS-TIs (S.O.). Next, we examine FSS-TIs (T.O.). In this setting, the model has no access to the source domain and thus does not explicitly account for domain shift. The results show that, even without source-domain training, FSS-TIs can still achieve satisfactory accuracy by fine-tuning with a very limited number of target-domain samples. This explains why FSS-TIs is insensitive to the severity of domain shift, as target-domain fine-tuning effectively bridges the domain gap and alleviates the impact of severe domain shifts. However, this also raises another question: if satisfactory performance can be achieved without source-domain training, does this imply that source-domain training is unnecessary? Referring again to Table~\ref{TAB_ABS-SADS}, we find that the standard deviation of FSS-TIs (T.O.) is significantly higher than that of FSS-TIs. This suggests that although target-domain fine-tuning dominates the accuracy improvement, source-domain training is crucial for ensuring model stability. Therefore, source-domain training remains indispensable.

Furthermore, since source-domain training does not significantly improve the overall accuracy of FSS-TIs, does this imply that the proposed TTIs fails to enhance the CD-FSS performance during source-domain training? To answer this question, we further report the performance of the model without TTIs, denoted as FSS-TIs$-$TTIs. We observe that, regardless of the number of support images, the accuracy of the model trained only on the source domain without TTIs (FSS-TIs$-$TTIs (S.O.)) drops significantly. This indicates that, even in the source-domain training stage, TTIs still substantially improves the CD-FSS capability of the model. Moreover, the accuracy of the model fine-tuned only on the target domain without TTIs (FSS-TIs$-$TTIs (T.O.)) is also significantly lower than that of FSS-TIs (T.O.). This further demonstrates the dominant role of TTIs.

\subsubsection{Sensitivity to Support Sizes}

While Table~\ref{TAB_ABS-SADS} presents the segmentation accuracies under different levels of domain-shift severity, it also reports the segmentation performance of different FSS-TIs variants as the number of support images gradually increases. From the accuracy comparison in Table~\ref{TAB_ABS-SADS}, it can be observed that the accuracy of FSS-TIs consistently improves as the number of support images increases. Moreover, FSS-TIs (S.O.), which is trained only on the source domain, exhibits a trend of fast improvement followed by slower improvement, that is, the gain from 1-shot to 5-shot is relatively large, while the gain from 5-shot to 10-shot is relatively small. In contrast, for FSS-TIs (T.O.), which performs only target-domain fine-tuning, the accuracy improves substantially each time the number of support images increases. FSS-TIs shows a similar trend to FSS-TIs (T.O.). This phenomenon is not difficult to understand, because the number of training samples of each category involved in target-domain fine-tuning is constrained to be equal to the number of support images in this paper to meet the real-world requirements. As the number of support images increases, more target-domain images and corresponding label maps become available for fine-tuning, thereby providing more supervisory information, which is beneficial for improving accuracy.

\subsubsection{Discovery of Domain-Agnostic Features}
Taking the cross-domain task of PASCAL VOC $\rightarrow$ DeepGlobe as an example, this paper verifies the domain-agnostic feature mining ability of TTIs and the category discriminability of the domain-agnostic feature space. To evaluate the effectiveness of domain-agnostic feature mining, all images from the source-domain dataset PASCAL VOC are fed into FSS-ITs, where both the raw features (produced by the feature extraction backbone) and the transformed features (obtained after the transformation of TTIs) are generated. For each image, the mean vector of its features is computed. Then, these mean vectors collectively form the feature space of the entire source domain. For the target-domain dataset DeepGlobe, we construct the feature space in the same manner. Since this feature space is high-dimensional (512 dimensions), we project the high-dimensional features into a two-dimensional space while preserving as much information as possible using Principal Component Analysis (PCA). The overlap between the source- and target-domain feature spaces is then examined, which serves as the basis for evaluating the domain-agnostic feature mining capability.

\begin{figure}[!t]
\centering
\includegraphics[width=3.5in]{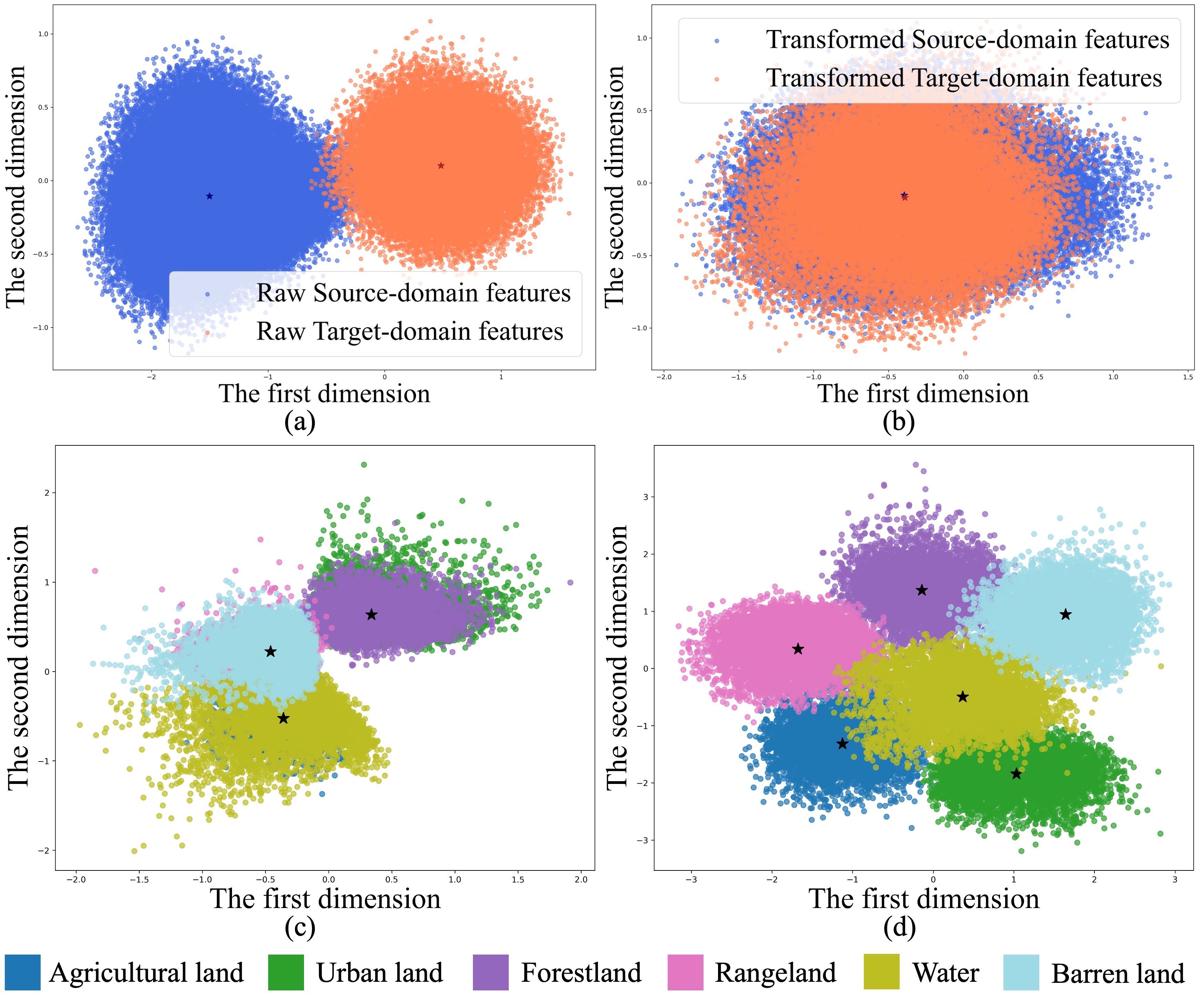}
\caption{The feature distributions of source and target domains, and all the novel categories in target domain.}
\label{FIG_ABS-DDAFS}
\end{figure}

Figs. \ref{FIG_ABS-DDAFS} (a–b) present the visualization results of the feature spaces from the source and target domains. Specifically, Fig. \ref{FIG_ABS-DDAFS} (a) presents the raw feature space, where the source- and target-domain feature spaces do not overlap and exhibit a clear deviation, indicating a significant domain shift between PASCAL VOC and DeepGlobe. However, after transformation by TTIs, as shown in Fig. \ref{FIG_ABS-DDAFS} (b), the two feature spaces overlap, suggesting that the domain shift has been effectively suppressed and that TTIs has successfully identified a domain-agnostic feature space.

Figs. \ref{FIG_ABS-DDAFS} (c–d) display the visualization results of the feature distributions for the novel categories in the target-domain dataset DeepGlobe. To visualize the feature distribution of each novel category, we adopt a similar approach to that in Figs. \ref{FIG_ABS-DDAFS} (a–b). The difference is that we compute a mean vector for the features of each novel category in every image, and these mean vectors collectively form the feature distribution of each novel category in the dataset. In Fig. \ref{FIG_ABS-DDAFS} (c), which depicts the raw feature space, the feature distributions of the novel categories largely overlap. Although the overlap among the feature distributions of {\it water}, {\it barren land} and {\it forestland} is relatively limited, all of these categories still exhibit complete overlap with the distributions of other categories. In contrast, Fig. \ref{FIG_ABS-DDAFS} (d) presents the feature space after transformation by TTIs, where the feature distributions of the novel categories are separated. While a small amount of confusion remains in the boundary regions of the distributions, the category-aware discriminability of the features is significantly improved compared with the raw feature space. 

\section{Conclusion}
This paper proposes a novel CD-FSS method, FSS-TIs, which provides an all-in-one, easy-to-use domain adaptation module TTIs. TTIs leverages ODEs to integrate the Fast Fourier Transform and random spectral perturbation, and further formulates feature transformations and augmentations along the time sequence as a parameter learning process embedded within the ODEs. This design facilitates the transformation of domain-specific features into a domain-agnostic feature space, thereby endowing the model with cross-domain generalization capability. In addition, this study imposes stricter constraints on the use of support images in target-domain fine-tuning, making the fine-tuning and testing stages more consistent with real-world scenarios. 

The experiments introduce nine datasets from distinctly different domains for cross-domain testing, demonstrating the advantages of FSS-TIs over existing methods. Extensive ablation studies validate the positive impact of each component in FSS-TIs. By separating source-domain training and target-domain fine-tuning for in-depth analysis, we further demonstrate the influence of domain-shift severity and the number of support samples on FSS-TIs. Combined with feature visualization, the results verify the role of TTIs in exploring the domain-agnostic feature space, as well as the discriminability of novel categories within that space.

At present, this study introduces the initial value problem of ODEs to enhance the CD-FSS capability, and the segmentation performance has been verified to be superior. In future work, we will further explore the role of other differential equations in cross-domain tasks, seeking theoretically more reliable and practically more effective modeling approaches.

\bibliographystyle{IEEEtran}
\bibliography{CDFSS_NiH}

\vfill

\end{document}